\documentclass[letterpaper, 10 pt, journal, twoside]{IEEEtran}

\IEEEoverridecommandlockouts                              

\usepackage{amsmath,epsfig}
\usepackage{subfig} 
\usepackage{hhline}
\usepackage{graphics}
\usepackage{bbm} 
\usepackage{textcomp} 

\usepackage{multirow} 
\usepackage{multicol} 
\usepackage{hhline}
\usepackage{amssymb} 

\usepackage[labelsep=period]{caption}

\def\etal{\emph{et al}.\ }

\makeatletter
\let\old@ps@headings\ps@headings
\let\old@ps@IEEEtitlepagestyle\ps@IEEEtitlepagestyle
\def\confheader#1{%
  \def\ps@headings{%
    \old@ps@headings%
    \def\@oddhead{\strut\hfill#1\hfill\strut}%
    \def\@evenhead{\strut\hfill#1\hfill\strut}%
  }%
  \def\ps@IEEEtitlepagestyle{%
    \old@ps@IEEEtitlepagestyle%
    \def\@oddhead{\strut\hfill#1\hfill\strut}%
    \def\@evenhead{\strut\hfill#1\hfill\strut}%
  }%
  \ps@headings%
}
\makeatother

\confheader{
\begin{minipage}{\textwidth}
  \centering
  This is the author's version of an article that has been accepted to IEEE Robotics and Automation Letters 2018. Changes \\ were made to this version  
  by the publisher prior to publication. The final version of record is available at http://dx.doi.org/10.1109/LRA.2018.2792681.
  \end{minipage}
}

\makeatletter
\def\ps@IEEEtitlepagestyle{
  \def\@oddfoot{\mycopyrightnotice}
  \def\@evenfoot{}
}
\def\mycopyrightnotice{
  {\footnotesize
  \begin{minipage}{\textwidth}
  \centering
  Copyright~\copyright~2017 IEEE. Personal use of this material is permitted. However, permission to use this  \\ 
  material for any other purposes must be obtained from the IEEE by sending a request to pubs-permissions@ieee.org.
  \end{minipage}
  }
}

\title{
Noise-resistant Deep Learning for Object Classification in 3D Point Clouds Using a Point Pair Descriptor
}

\author{Dmytro Bobkov$^{1}$, Sili Chen$^{2}$, Ruiqing Jian$^{1}$, Muhammad Z. Iqbal$^{1}$, Eckehard Steinbach$^{1}$%
\thanks{$^{1}$Dmytro Bobkov, Ruiqing Jian, Muhammad Z. Iqbal and Eckehard Steinbach are with the Chair of Media Technology, Technical University of Munich, 80333 Munich, Germany. Contact email: {\tt\small dmytro.bobkov@tum.de}}
\thanks{$^{2}$Sili Chen is with the Augmented Reality Lab, Baidu Inc., 100193, Beijing, China.}
}

\author{Dmytro Bobkov$^{1}$, Sili Chen$^{2}$, Ruiqing Jian$^{1}$, Muhammad Z. Iqbal$^{1}$, Eckehard Steinbach$^{1}$%
	\thanks{Manuscript received: September, 10, 2017; Revised November, 29, 2017; Accepted December, 22, 2017.}
	\thanks{This paper was recommended for publication by Editor Tamim Asfour upon evaluation of the Associate Editor and Reviewers' comments. Muhammad Z. Iqbal has been supported by a PhD scholarship provided by the Higher Education Commission (HEC) of Pakistan.}
	\thanks{$^{1}$Dmytro Bobkov, Ruiqing Jian, Muhammad Z. Iqbal and Eckehard Steinbach are with the Chair of Media Technology, Technical University of Munich, 80333 Munich, Germany
		{\tt\small dmytro.bobkov@tum.de}}%
	\thanks{$^{2} $Sili Chen is with the Augmented Reality Lab, Baidu Inc., 100193, Beijing, China
		{\tt\small sili.chen@foxmail.com}}%
	\thanks{Digital Object Identifier (DOI): see top of this page.}
}

\begin{document}

\maketitle

\begin{abstract}

Object retrieval and classification in point cloud data is challenged by noise, irregular sampling density and occlusion. To address this issue, we propose a point pair descriptor that is robust to noise and occlusion and achieves high retrieval accuracy. We further show how the proposed descriptor can be used in a 4D convolutional neural network for the task of object classification. We propose a novel 4D convolutional layer that is able to learn class-specific clusters in the descriptor histograms. Finally, we provide experimental validation on 3 benchmark datasets, which confirms the superiority of the proposed approach.

\end{abstract}


\section{Introduction}
\label{sec:intro}

\IEEEPARstart{O}{bject} retrieval and classification are important tasks in robotics. Of all 3D data representations, point clouds are closest to the output from LiDAR and depth sensors. Unfortunately, this representation is very challenging due to its irregular data structure and large data size. Because of this, many works first convert this representation into 3D voxel grids or multi-view rendered images. While convenient for processing, this pre-processing step introduces additional computational complexity and makes the resulting data representation unnecessarily voluminous. For this reason, we focus on approaches that work directly on point clouds. There are a number of handcrafted descriptor algorithms (\cite{Birdal2015}, \cite{Wohlkinger2011}, to name a few) designed for point cloud data that are able to produce a fixed size descriptor. Unfortunately, their description performance is limited for realistic data that suffer from high levels of noise and occlusion. These disturbances often occur in robotic applications, where an object cannot be scanned from all viewpoints due to time, energy and accessibility constraints.

Point pair-based global descriptors \cite{Birdal2015}, \cite{Wohlkinger2011}, \cite{Drost2010}, \cite{Wahl2003} achieve high performance for object matching and classification as they do not describe the geometry explicitly and instead use point pair functions (PPF). The descriptor values describing shape statistics are computed using PPFs on the point pairs. In this paper, similarly to \cite{Drost2010}, we quantize PPF values using a 4D histogram.

\begin{figure}
	\centering
	\parbox{0.99\linewidth}{
\includegraphics[width=0.99\linewidth]{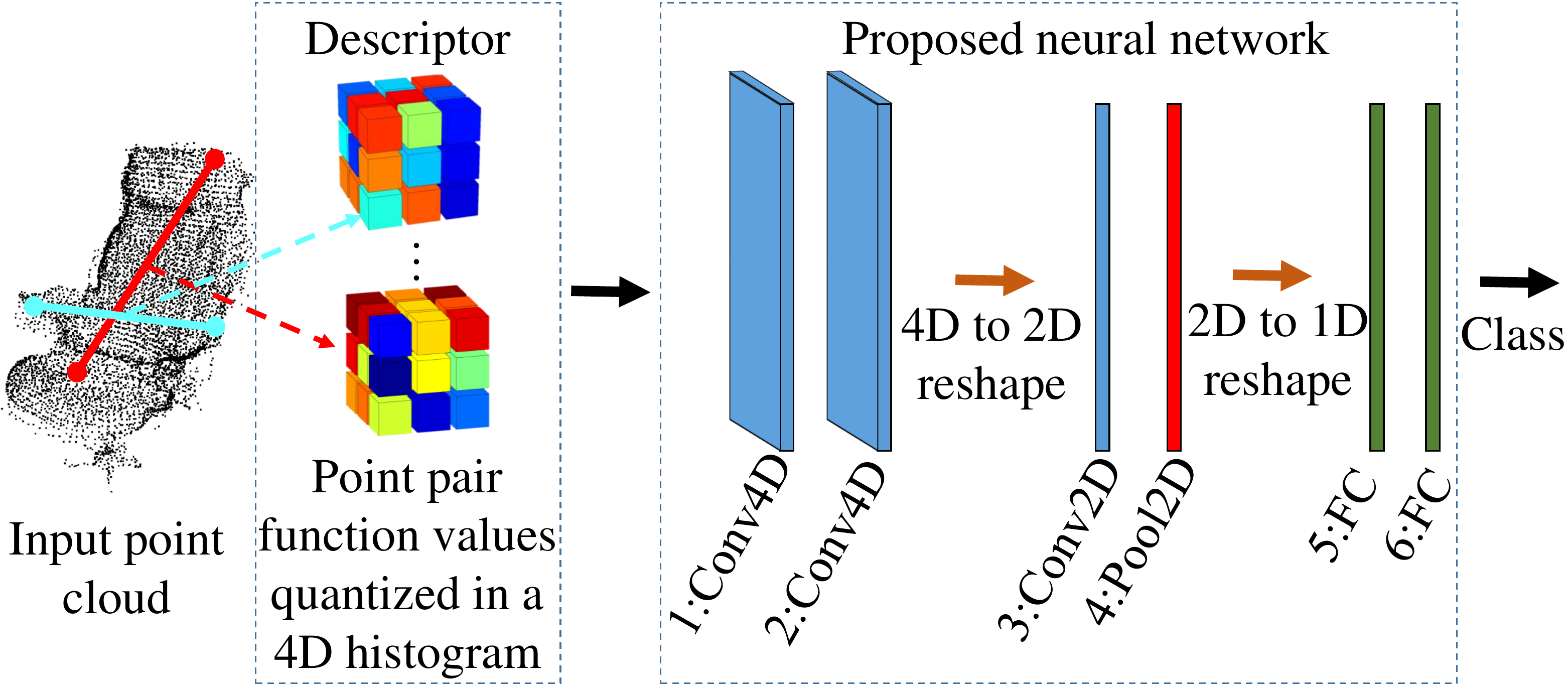}}
	\caption{Overview of the proposed object classification pipeline that is a combination of a novel handcrafted descriptor and a 4D convolutional neural network (CNN). For details on the network architecture and layer dimensions, see Fig.~\ref{fig:cnn_architecture} and Table~\ref{tab:cnn_architecture}. Here, FC denotes a fully connected layer.}
	\label{fig:overview}
\end{figure}

Deep learning-based approaches have experienced significant interest over the last years for the task of 3D object classification. While a number of approaches convert the 3D data to a regular representation for the ease of processing, we argue that it introduces an unnecessary computational step. However, due to the irregular data structure of point clouds, it is not straightforward how to feed this representation into a neural network. In particular, a point cloud is an unstructured set of points, which is invariant to permutations of its members. Different permutations of the set members result in different input data for the network, which makes training of the neural network complex (also known as symmetrization problem). PointNet \cite{Qi2016} addresses this problem by using max-pooling. However, the classification performance remains relatively low for realistic datasets that are subject to noise and occlusion. We argue that this is because the network is not able to learn sufficiently invariant representations based solely on point sets. To address these limitations, we instead feed the 4D descriptor into the neural network (see Fig.~\ref{fig:overview}). For this, we employ a novel network architecture with 4D convolutional layers, which outperforms state-of-the-art deep learning-based approaches on benchmark datasets.

\textbf{Contributions} of this paper are as follows:
\begin{enumerate}
\item We present a novel 4D convolutional neural network architecture that takes a 4D descriptor as input and outperforms existing deep learning approaches on realistic point cloud datasets.
\item We design a handcrafted point pair function-based 4D descriptor that offers high robustness for realistic noisy point cloud data. The source code of the descriptor will be made publicly available\footnote{https://rebrand.ly/obj-desc}.
\end{enumerate}

\section{Related Work}

\textbf{Handcrafted point cloud descriptors}. Point cloud descriptors are typically divided into global and local descriptors. Local descriptors usually offer superior performance, but they come with high complexity as typically thousands of keypoints need to be computed and matched per object~\cite{Rusu2009}. Instead, global descriptors compute one descriptor per object, thus reducing the matching complexity. Some of the global descriptors use projection techniques \cite{Kasaei2016}, \cite{Sanchez2015}. The descriptor proposed by Kasaei \etal \cite{Kasaei2016} exhibits state-of-the-art accuracy and achieves low computational complexity. Sanchez \etal \cite{Sanchez2015} combine local and global descriptors into a mixed representation. Others employ PPFs, which are based on sampled point pairs \cite{Birdal2015}, \cite{Wohlkinger2011}, \cite{Drost2010}, \cite{Wahl2003}, \cite{Aldoma2012}, \cite{Furuya2015}. Such descriptors are more robust to noise and occlusion that often occur in real point cloud datasets. Unfortunately, their recognition performance is still not sufficient for many applications.

\textbf{Deep learning on point clouds}. Deep learning-based approaches have shown significant success over the last few years. However, due to the irregular structure of point cloud data and issues with symmetry, many approaches convert the raw point clouds into 3D voxel grids (\cite{Sedaghat2017} and \cite{Engelcke2017}, to name a few). Volumetric representations, however, are constrained by their resolution due to data sparsity and unnecessary convolutions over empty voxels. Some other approaches solve the problem using field probing neural networks \cite{Li2016}, but it still remains unclear how to apply such approach on a general problem. Multi-view CNNs achieve a good performance \cite{Su2015}, but yet require an additional step of view rendering. There are also a number of feature-based CNNs that first compute features based on 3D data and then feed these into a neural network. While they achieve good results \cite{Xie2016}, the presented features are only suitable for mesh structures and it is unclear how to extend them to point sets. Finally, there are a number of end-to-end deep learning methods working directly on point sets, such as \cite{Ravanbakhsh2017}, \cite{Qi2017}, \cite{Qi2016}. However, these are sensitive to noise and occlusion in real point cloud datasets. 

Due to limitations of related work, we review existing PPFs and propose the new ones. The quantized PPFs of the descriptor are then input into a 4D convolutional neural network for the task of object classification.

\section{Methodology}

\subsection{Handcrafted Descriptors}

Sampling-based PPFs were previously used for the construction of robust 3D descriptors by \cite{Birdal2015}, \cite{Wohlkinger2011}, \cite{Wahl2003}, \cite{Hinterstoisser2016}. Typically, point pairs and point triplets are randomly sampled from the point set. Based on the sampled pairs and triplets, the functions map them to the scalar values, which are then quantized into a histogram that describes the statistics of the shape. Such point sampling leads to certain randomness in the result, but also enhances robustness to noise and occlusion. A further advantage of this approach is its rotation-invariance.

We define a point pair function $f$ as the mapping of a point pair to a scalar value as follows:
\begin{equation}
f: (\mathbb{R}^{3}, \mathbb{G}) \times (\mathbb{R}^{3}, \mathbb{G}) \rightarrow \mathbb{R}^{1},
\end{equation}
with $\mathbb{R}^{3}$ being a Euclidean space and $\mathbb{G}$ denoting a manifold of surface normal orientations in 3D space. We employ the following \textbf{functions} $f_1$ to $f_4$ for our point pair-based descriptor:
\begin{enumerate}
\item Euclidean distance between the points $f_1$ \cite{Wahl2003}.
\item The maximum angle between the corresponding surface patches of the points and direction vector $d$ connecting the points $f_2$.
\item Normal distance between the points $f_3$.
\item Occupancy ratio along the line connecting the points $f_4$.
\end{enumerate}

\begin{figure}
	\centering
	\makebox{\parbox{0.6\linewidth}{
\includegraphics[width=0.99\linewidth]{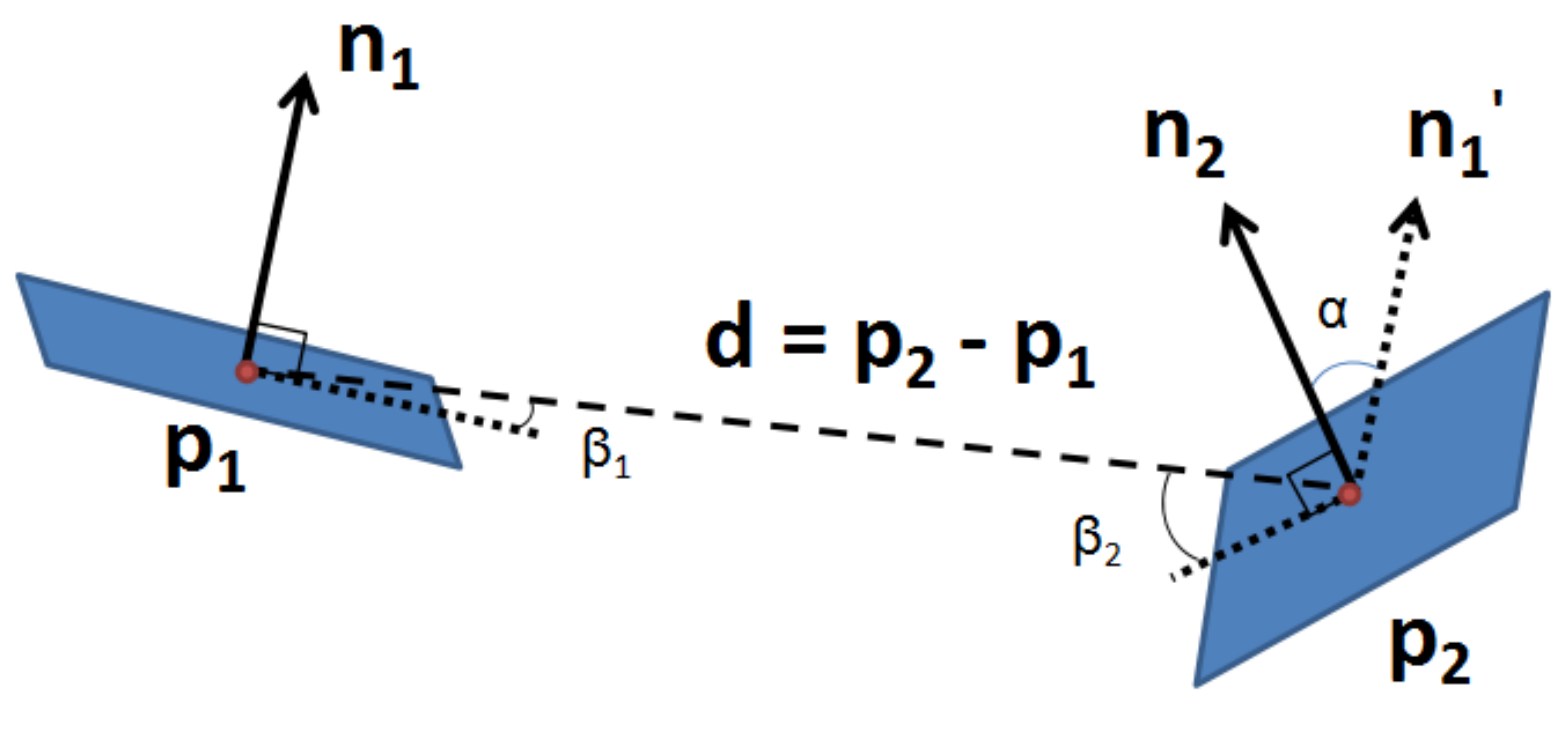}}}
	\caption{Illustration of the points $p_1$, $p_2$, their normal vectors $n_1$, $n_2$ and the Euclidean distance $f_1=\| d \|_2$.
	}
	\label{fig:feature_illustration}
\end{figure}

\textbf{Euclidean distance}. The function value $f_1$ is the Euclidean distance between two points $p_1$ and $p_2$:
\begin{equation}
f_1(p_1, p_2) = \| p_2 - p_1 \|_2.
\end{equation}
The statistics of the distances between point pairs can represent both the geometry and the size of the object (see $d$ in  Fig.~\ref{fig:feature_illustration}).

\begin{figure}
	\centering
	\makebox{\parbox{0.8\linewidth}{	\centering
\includegraphics[width=0.99\linewidth]{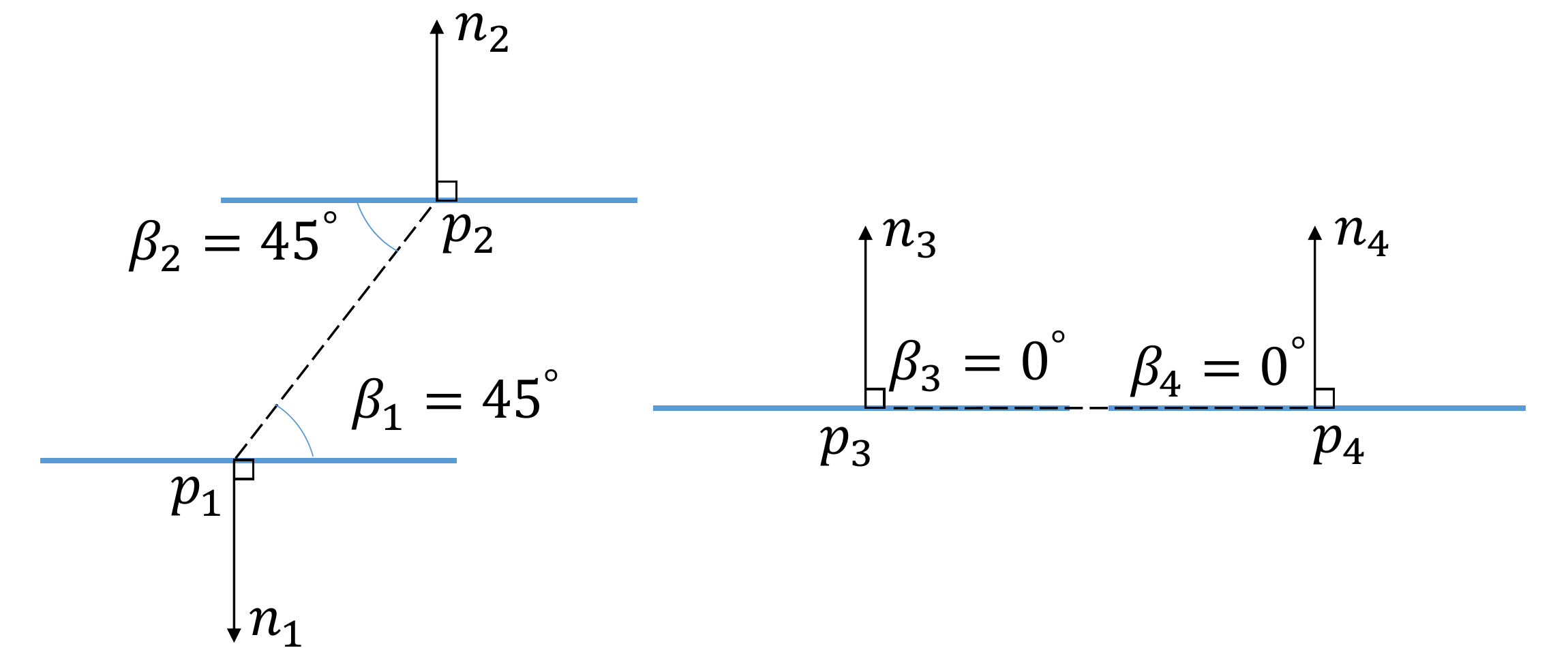}}}
	\caption{Illustration of the case with point pairs $(p_1, p_2)$ (left) and $(p_3, p_4)$ (right) that have similar Euclidean ($f_1$) and normal distances ($f_3$), but still describe significantly different shapes. The maximum angle between patches and direction $f_2$ is an important feature for such case.}
	\label{fig:feature3_cases}
\end{figure}

\textbf{The maximum surface angle}. Function value $f_2$ describes the underlying patch orientation with respect to the line connecting both points $d$. It is defined as follows:
\begin{equation}
f_2(p_1, n_1, p_2, n_2) = \max(\beta_1 , \beta_2),
\end{equation}
with $\beta_1 = \arccos(n_1 \cdot d) - \pi/2$ and $\beta_2 =  \arccos(n_2 \cdot d) - \pi/2$ being the angles between vector $d$ and the tangent patches of points $p_1$ and $p_2$, respectively (see Fig.~\ref{fig:feature_illustration}). The direction vector is computed as follows: $d=p_2-p_1$. Function $f_2$ is important in the cases, when the Euclidean distance $f_1$ and normal distance $f_3$ are not descriptive enough. Refer to Fig.~\ref{fig:feature3_cases} for illustration. Here, the point pairs in the left and right have the same Euclidean distance and the same normal distance. In contrast, function $f_2$ has significantly different values for both pairs, hence it provides descriptive information on the geometry. Although $\beta_1$ and $\beta_2$ may take on different values and, hence, provide an informative surface description, we choose only one value. Thus, we strike a trade-off between compactness and accuracy. We use the maximum operation as compared to minimum or average because we have observed that it is much more descriptive for noisy datasets.

\textbf{Normal distance}. Function value $f_3$ describes the similarity of surface orientations of the point neighborhoods, which is defined as follows:
\begin{equation}
f_3(n_1,n_2) = \arccos(\lvert n_1 \cdot n_2 \lvert),
\end{equation}
which lies in the range from $0$ to $\pi$. We take the absolute value of the dot product, because we want to eliminate the influence of the viewpoint, which can be unreliable in multi-view point clouds \cite{Bobkov2017}. 

\begin{figure}
	\centering
	\makebox{\parbox{0.38\linewidth}{ \centering
\includegraphics[width=0.99\linewidth]{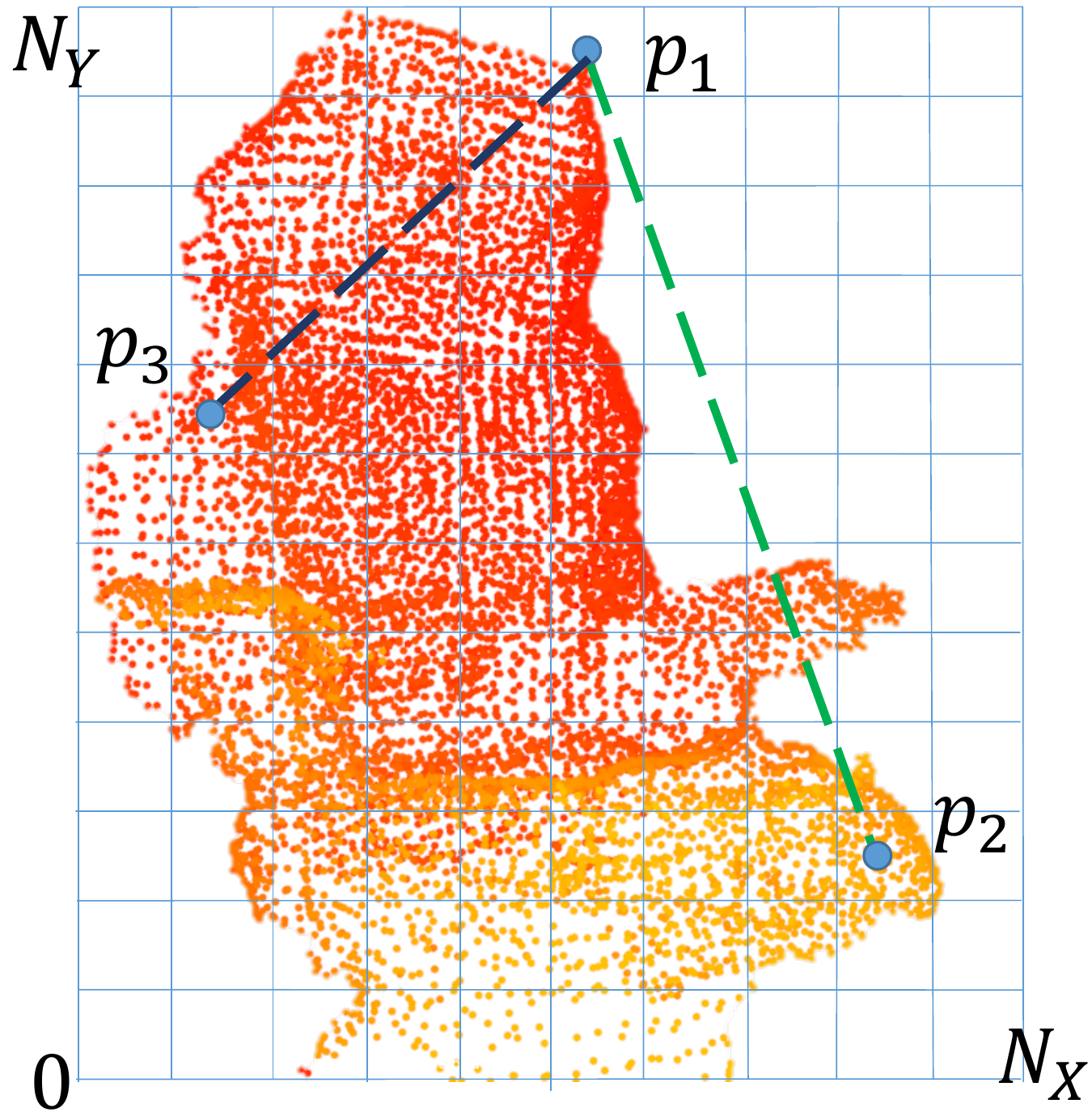}}}
	\caption{2D illustration of the grid used for performing voxel occupancy checks for the voxels lying along the line (dashed line) connecting a given point pair $p_1$ and $p_2$ and another point pair $p_1$ and $p_3$.}
	\label{fig:voxel_grid}
\end{figure}

\textbf{Occupancy ratio}. The value $f_4$ describes the object geometry. In particular, we perform voxelization of the object volume using a voxel grid of dimensions $N_x \times N_y \times N_z$ to enable fast lookup and occupancy check computations \cite{Wohlkinger2011}. We set $N_x=N_y=N_z=64$. The value of $f_4$ is defined as follows:
\begin{equation}
f_4(P, p_1, p_2) = \frac{N_{occ}}{N_{total}},
\end{equation}
with $P \in \mathbb{R}^{3}$ being the set of points in the considered point cloud, $N_{occ}$ is the number of occupied voxels intersected by the 3D line connecting two points $p_1$ and $p_2$, and $N_{total}$ is the total number of voxels intersected by the line (see Fig.~\ref{fig:voxel_grid}). We classify the voxel as occupied if at least one point is contained inside. This is because the point density can vary significantly in indoor point clouds, therefore such conservative value provides higher robustness. This function describes the global object geometry, because the voxel grid occupancy is computed based on all points. For the example with $p_1$ and $p_2$ in Fig.~\ref{fig:voxel_grid}, $N_{occ}=8$, $N_{total}=13$, leading to $f_4=0.615$. In contrast, for point pair $(p_1, p_3)$,  $f_4=1$ as all intersected voxels are occupied.

\subsection{Feature Statistics}

\begin{figure}
	\centering
	\makebox{\parbox{0.9\linewidth}{ \centering
\includegraphics[width=0.99\linewidth]{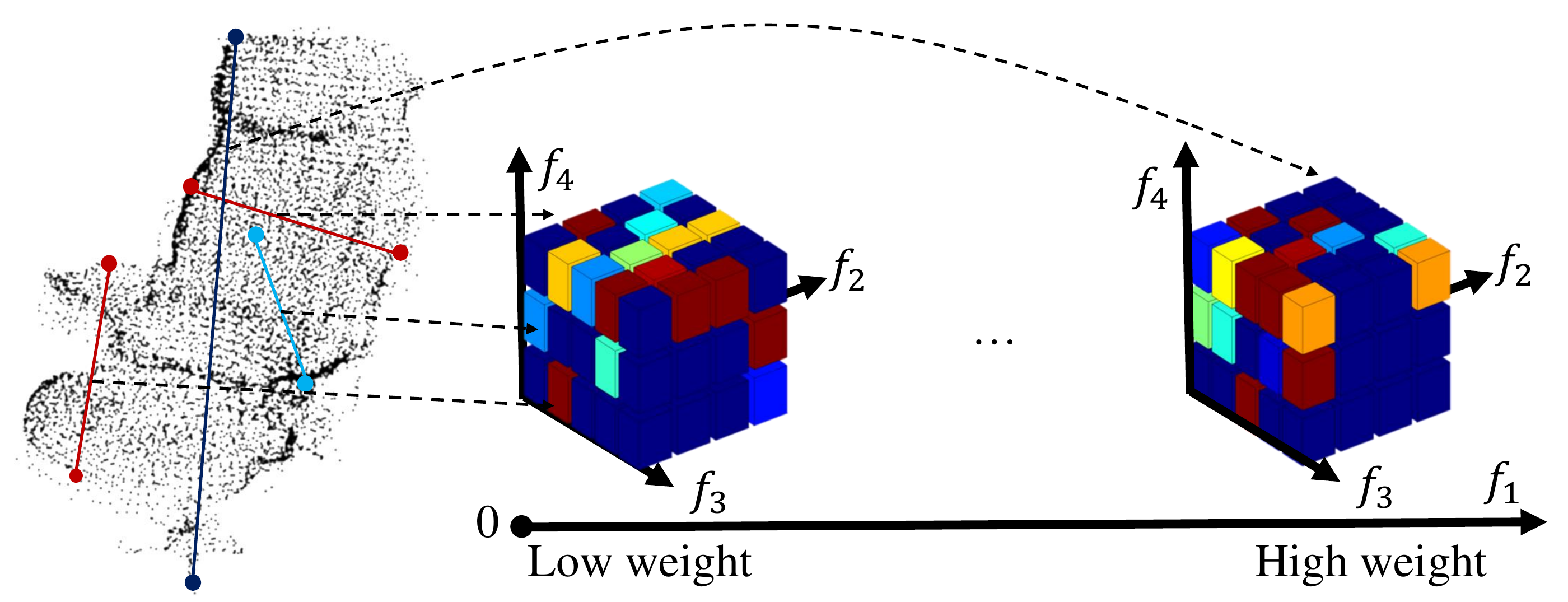}}}
	\caption{4D histogram that is used to discretize the aggregated counts of sampled PPF values $w_{i,j,k,l}$ into a descriptor. Blue color denotes the bins with low number of counts, whereas red corresponds to high.}
	\label{fig:4d_histogram}
\end{figure}

We draw $20,000$ point pairs at random from the point set as we observe that this is sufficient to describe complex shapes. After the PPF values are computed for these pairs, they need to be aggregated into a descriptor histogram. Many approaches (among others \cite{Wohlkinger2011} and \cite{Aldoma2012}) assume that the different function values are uncorrelated with each other. Therefore, these function values have been discretized into bins and concatenated into a 1D histogram. We have observed that aggregation into a 1D histogram leads to significant loss of performance because information on co-occurrences of different function values is neglected. To avoid loss of information on 4D co-occurrences, similar to \cite{Wahl2003}, we instead build a 4D histogram of function value occurrences (shown in {Fig.~\ref{fig:4d_histogram}}). It can be expressed as:
\begin{equation}
F = (f_1, f_2, f_3, f_4).
\end{equation}
We denote the descriptor as Enhanced Point Pair Functions (EPPF). Clearly, a straightforward extension into a 4D histogram would result in an exponential increase of computational complexity \cite{Drost2010, Wahl2003}. We observe that not all function values are equally informative for the description of the object geometry, hence a different number of bins has to be chosen for different dimensions. In the later section \ref{sec:obj_retrieval} we provide an experimental study of the feature contribution to the overall performance.

In particular, function $f_1$ helps to distinguish objects of different sizes, therefore we choose a relatively large number of bins $N_{f_1} = 20$. Contrary to scale-variant approaches, we do not scale every object to fit into a unit sphere, as we observe that in indoor environments the size of the object provides quite important information. For example, \textit{monitor} and \textit{whiteboard} can have similar geometric shape, but different dimensions. To preserve this information, we make our descriptor scale-variant and scale all objects so that the largest one fits into a unit cube. For $f_2$, we observe that its descriptive ability is relatively low for noisy and occluded data, therefore we choose a relatively small number of bins $N_{f_2}=4$. Furthermore, for $f_3$ we set it to a slightly larger number $N_{f_3} = 5$.  Finally, for $f_4$, we observe that $N_{f_4}=3$ is sufficient. We have experimented with larger numbers of bins and have noticed no significant performance improvement. Thus, we strike a trade-off between complexity and accuracy of the descriptor.

According to our observation, point pairs with larger Euclidean distances usually have higher discriminative power as compared to those with smaller Euclidean distances. This is because every object has point pairs with small distance, but only certain objects have point pairs with larger distances.  Furthermore, to suppress the influence of noise in low distance regions and enhance the difference in high distance regions for a better discrimination, we use the following bin weighting factor for the bin located at index $i,j,k,l$:
\begin{equation}
\alpha_i = \ln(i/N_{f_1} + c),
\end{equation}
where $i$ is the index of the Euclidean bin, and $\alpha_i$ is used to compute the bin weight as follows: $w^{new}_{i,j,k,l}=\alpha_i \cdot w_{i,j,k,l}$ (see $w_{i,j,k,l}$ in Fig.~\ref{fig:4d_histogram}). $c$ is a constant greater than 1, which is used to guarantee that weights are positive and mitigate noise for point pairs with smaller Euclidean distances. Based on the experimental validation, we set $c = 1.2$. Finally, every weighted descriptor histogram is normalized. The total number of bins of the resulting 4D histogram is $N=1200$. For matching, we employ the symmetrized form of Kullback-Leibler (KL) divergence as a distance metric between two histograms:
\begin{equation}
d(W1, W2) = \sum_{i=1}^N (W1_i - W2_i) \ln \frac{W1_i}{W2_i},
\end{equation}
where $W1$ and $W2$ are histogram counts for object 1 and 2, respectively. Similarly to {\cite{Wahl2003}}, we set all zero bins of a histogram to a common minimum value that is twice smaller than the smallest observed bin value in this dataset. We observed that KL outperforms L1 and L2-distance metrics.

\subsection{4D Deep Learning Architecture}

\begin{figure*}
	\centering
	\makebox{\parbox{0.9\linewidth}{ \centering
\includegraphics[width=0.99\linewidth]{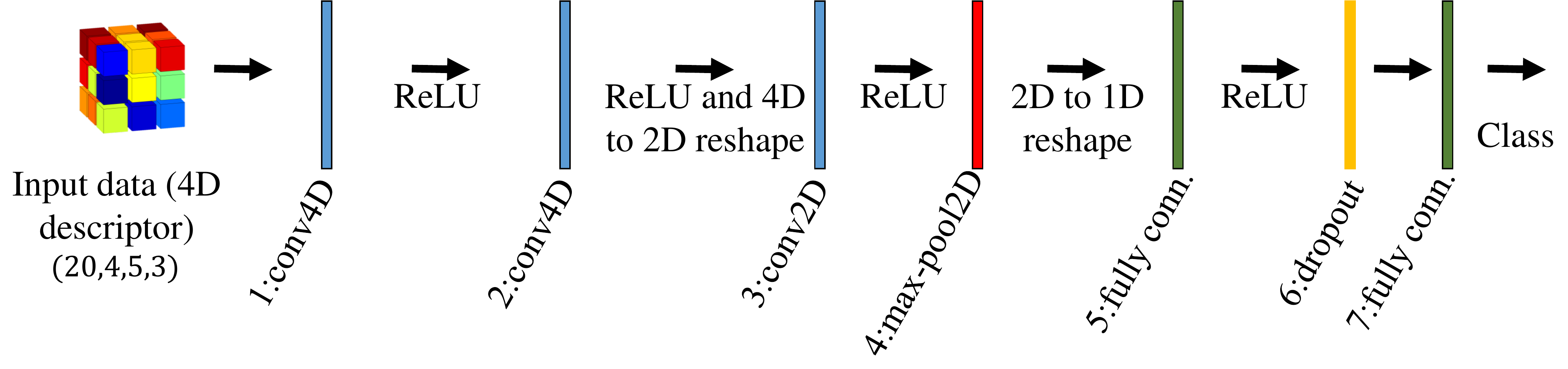}}}
	\caption{ Architecture of the proposed 4D neural network. See Table \ref{tab:cnn_architecture} for more details on the dimensions.}
	\label{fig:cnn_architecture}
\end{figure*}

The previously computed 4D descriptor is rotation-invariant, which resolves the issue of symmetry of point sets in neural networks. Hence, it is possible to feed this representation into a neural network for the task of object classification. To preserve information about 4D co-occurrences of the function values, we use 4D convolution for the first two layers of the neural network (see Fig.~\ref{fig:cnn_architecture}). 4D convolution has already been successfully applied for the task of material recognition by Wang \etal \cite{Wang2016}, where it outperformed other architectures. In particular, we use 4D convolutional blocks in the first and second layers, respectively. Details on the dimensions are given in  Table~\ref{tab:cnn_architecture}. Afterwards, the resulting responses are reshaped into a 2D structure and fed into a 2D layer block. Furthermore, 2D max-pooling is performed. This is followed by reshaping from 2D to 1D representation, which is input into a fully connected layer. Afterwards, to achieve regularization and enhance the generalization property of the network, we employ a dropout layer, which is followed by a fully connected layer. At the output of the network the class prediction for the object is provided.

\begin{table}
\centering
	\caption{Layer dimensions for 2D, 3D and 4D-variants of the network. $N_f$ denotes the number of filters.}
	\label{tab:cnn_architecture}
	\begin{tabular}{| c | c | c | c | c |}
	\hline
	Layer & 2D network & 3D network & 4D network & $N_{f}$ \\
	\hline
	Input & $(40,30)$ & $(20,4,15)$ & $(20,4,5,3)$ & - \\
	1 (conv.) & 2D:$(5,5,1)$ & 3D:$(5,5,1,1)$ & 4D:$(5,2,2,1,1)$ & 32 \\
	2 (conv.) & 2D:$(5,5,32)$ & 3D:$(5,5,1,32)$ & 4D:$(5,2,2,1,32)$ & 64 \\
	3 (conv.) & 2D:$(5,5,64)$ & 3D:$(5,5,1,64)$ & 2D:$(5,5,64)$ & 48 \\ \hline
	4 & \multicolumn{3}{c|}{max-pool: (2,2)} & 1 \\
	5 & \multicolumn{3}{c|}{fully connected (192,1024)} & 1 \\
	6 & \multicolumn{3}{c|}{dropout 0.5} & 1 \\
	7 & \multicolumn{3}{c|}{fully connected: (1024,$N_{classes}$)} & 1 \\ \hline
	\end{tabular}
\end{table}

For comparison, we also design 2D- and 3D-convolution-based networks for object classification. For fair evaluation, we choose the dimensions so that the number of parameters of all three networks is comparable to each other. The dimensions of the single layers are given in Table~{\ref{tab:cnn_architecture}}. Thus, for the 2D-variant of the network, the input 4D descriptor is first reshaped into 2D with dimensions $(40,30)$ and then processed with three 2D convolution layers. For a 3D-variant of the network, the input 4D descriptor is reshaped into 3D with dimensions $(20,4,15)$ and then processed with three 3D convolutional layers. The number of filters remains the same for all three networks. We use stride value of 1.

\section{EXPERIMENTAL RESULTS}
\label{sec:results}

For experimental comparison with state-of-the-art approaches, we choose OUR-CVFH \cite{Aldoma2012}, ESF \cite{Wohlkinger2011} and Wahl \etal \cite{Wahl2003}, as the ones that constantly perform well across various benchmarks. For the first two, we use the implementations provided in the Point Cloud Library 1.8 \cite{Aldoma2012PCL}. For Wahl \emph{et al}, we use our own implementation in C++. We further perform fine-tuning of the descriptor parameters to obtain optimal performance. For comparison with deep learning approaches, we use PointNet \cite{Qi2016}, as it is the only approach up-to-date that is able to directly work on 3D point sets without additional operations of multi-view projection \cite{Su2015} or voxelization \cite{Wu2015}. We use its implementation provided by the authors. 

The proposed descriptor EPPF has a larger number of bins as compared to OUR-CVFH and ESF, which raises the question, whether a larger number of bins has an impact on performance. To answer this question, we additionally evaluate the version of the EPPF descriptor with fewer bins. We choose its number so that it is comparable to the other descriptors. In particular, we set $N_{f_1}=15$, $N_{f_2}=3$, $N_{f_3}=4$, $N_{f_4}=3$. This results in a total number of bins of $N=540$: compare to $640$ in ESF and $308$ in OUR-CVFH. We refer to this descriptor as "EPPF short" in the following. As metrics for evaluation, we choose total accuracy that is accuracy value divided by the total number of objects, whereas mean accuracy and recall are averaged over the classes.

\subsection{Datasets}

For evaluation, we use the Stanford point cloud dataset \cite{Armeni2016}, ScanNet CAD dataset \cite{Dai2017} and ModelNet40 CAD dataset \cite{Wu2015}. These are the most recent and largest datasets of indoor objects that can be used for evaluation.

\textbf{Stanford dataset}. The Stanford dataset in \cite{Armeni2016} contains RGB and depth images and has been captured in 6 office areas within 3 different buildings, using structured-light sensors during a $360^{\circ}$ rotation at each scanning location. Due to sensor noise and limited scanning time, point density significantly varies throughout the scene. Furthermore, there is a high level of occlusion. The authors \cite{Armeni2016} propose a training/testing split according to buildings. We cannot use this split, because in this case some objects never occur in the testing or training sets. This would make the evaluation of object classification less meaningful, therefore we derive our own $60/40$ split. Here, $60\%$ of object instances per category make the training set and the remaining $40\%$ represent the testing set. We omit the category \textit{clutter}, as it contains different categories. Furthermore, we also skip architectural elements with a high level of planarity, such as \textit{floor}, \textit{ceiling} and \textit{wall}, as they can easily be classified using normal direction. The presence of these objects would make the object classification task unnecessarily complex. Thus, we have 10 classes in total with $3,735$ objects. 

\textbf{ScanNet dataset}. The ScanNet dataset \cite{Dai2017} is a large-scale CAD dataset containing semantic annotation of indoor scenes. It contains high level of occlusion and noise, as it is collected using a commodity RGB-D sensor in a low-cost sensor setup. For classification we employ the training/testing split specified by the authors \cite{Dai2017}. We use the list of categories that are compatible with ShapeNet 55 dataset and mentioned in \cite{Dai2017}. To avoid an unbalanced training set, we remove category \textit{laptop}, as it contains only $18$ instances (as compared to the others with at least $50$ instances). Thus, the used dataset contains $14$ categories: \textit{basket, bathtub, bed, cabinet, chair, keyboard, lamp, microwave, pillow, printer, shelf, stove, table and tv}. This results in $5,203$ objects in the training set and $1,699$ objects in the testing set.

\textbf{ModelNet40 dataset}. The ModelNet40 dataset \cite{Wu2015} is a large-scale CAD model dataset. The CAD models have been manually cleaned, thus containing practically no noise or occlusion. There are $12,311$ CAD models from $40$ categories, split into $9,843$ for training and $2,468$ for testing.

ModelNet40 and ScanNet datasets contain mesh models, which need to be converted into a point cloud representation. For this, we use the mesh sampling approach from the Point Cloud Library \cite{Aldoma2012PCL} with a resolution of 1 cm. Because EPPF, Wahl \etal and OUR-CVFH require normal information, we further perform normal estimation using the method of Boulch and Marlet~\cite{Boulch2012}.

\subsection{Object Retrieval using Handcrafted Descriptors}
\label{sec:obj_retrieval}

We perform leave-one-out cross-validation by querying every object in the dataset against the other objects to find the closest match. When the closest match is of the same category as the query object, we consider it as a correct retrieval, and incorrect otherwise. Because the ESF, Wahl \etal and EPPF descriptors contain the step of random sampling of point pairs from the point set, there are variations in performance as each time different pairs are chosen. To mitigate this, we repeat experiments ten times and record the mean and the standard deviation. The retrieval performance is given in Table~\ref{tab:handcrafted}.

\begin{table*} 
\small
	\centering
	\caption{ Retrieval performance of the handcrafted descriptors. The mean value is given in the corresponding column, while the standard deviation is given in brackets. Best performance is shown in bold.}
		\label{tab:handcrafted}
	\begin{tabular}{| c | c | c | c | c | c | c |}
	\hline 
	\multirow{3}{*}{Dataset} & \multirow{2}{*}{Metric} & \multicolumn{5}{c|}{Descriptor} \\ \cline{3-7}
	& & OUR-CVFH \cite{Aldoma2012} 	& ESF \cite{Wohlkinger2011} & Wahl \cite{Wahl2003} & EPPF Short & EPPF   \\ \cline{2-7}
	 & $N_{bins}$ & 308 & 640 & 625 & 540 & 1200 \\ \hline
	\multirow{4}{*}{Stanford \cite{Armeni2016}}	& Total accuracy (\%) & 62.79 & 71.34 ($\pm$0.82) & 75.13 ($\pm$0.35) & 77.26 ($\pm$0.40) & \textbf{80.18} ($\pm$0.40)  \\
	& Mean accuracy (\%) & 42.91 & 54.54 ($\pm$1.10) & 57.00 ($\pm$1.26) & 60.53 ($\pm$1.21)	& \textbf{64.01} ($\pm$0.66) \\
	& Mean recall (\%)	& 49.90 & 52.28 ($\pm$1.01) & 57.45 ($\pm$2.57)	& 60.16 ($\pm$2.61)	& \textbf{64.58} ($\pm$0.59)  \\
	& F1-score 	& 0.437 & 0.530 ($\pm$0.011) & 0.567($\pm$0.017) & 0.601 ($\pm$0.017) & \textbf{0.640} ($\pm$0.006)  \\         \hline
	    
	\multirow{4}{*}{ScanNet \cite{Dai2017}} & Total accuracy (\%) & 56.23 & 53.41 ($\pm$0.60) & 63.72 ($\pm$0.32) & 63.49 ($\pm$0.20) & \textbf{65.29} ($\pm$0.39)  \\
	& Mean accuracy (\%) & 39.83 & 33.69 ($\pm$0.82) & \textbf{45.40} ($\pm$0.65) & 42.02($\pm$0.49) & 44.95 ($\pm$0.69)  \\
	& Mean recall (\%)	& 38.21 & 32.72 ($\pm$0.98) & 45.94 ($\pm$0.46)	& 45.17 ($\pm$0.80)	& \textbf{47.54} ($\pm$1.00)  \\
	& F1-score 	& 0.382 & 0.327 ($\pm$0.008) & 0.444($\pm$0.005) & 0.430 ($\pm$0.006) & \textbf{0.457} ($\pm$0.008)  \\ \hline    
	        
	\multirow{4}{*}{M40 \cite{Wu2015}} & Total accuracy (\%)	& 53.22 & 65.87 ($\pm$0.37) & \textbf{74.41} ($\pm$0.24)	& 73.00 ($\pm$0.21)	& 73.68 ($\pm$0.21)  \\
	& Mean accuracy (\%) & 46.43 & 58.91 ($\pm$0.51) & \textbf{67.50} ($\pm$0.31)	& 65.79 ($\pm$0.26) & 66.43 ($\pm$0.31)  \\
	& Mean recall (\%)	& 49.26	& 59.96 ($\pm$0.68) & \textbf{70.33} ($\pm$0.33) & 69.12 ($\pm$0.36) & 69.79 ($\pm$0.30)  \\
	& F1-score & 0.465 & 0.588 ($\pm$0.005) & \textbf{0.680} ($\pm$0.003) & 0.666 ($\pm$0.003) & 0.671 ($\pm$0.003)  \\         \hline     
	        
	\end{tabular}
\end{table*}

One can observe in Table~\ref{tab:handcrafted} that the proposed EPPF descriptor (in full and short versions) outperforms ESF and OUR-CVFH on all datasets. Furthermore, the EPPF descriptor outperforms the Wahl descriptor on the Stanford and ScanNet datasets, but shows comparable performance on the M40 dataset. This is because of the low level of noise in this dataset. The PPFs employed in the Wahl descriptor are less robust to high levels of noise, but with lower noise levels can provide higher descriptive ability, as compared to the EPPF descriptor. Notably, there is a big difference in total and mean accuracy values for all descriptors. This is because the datasets are unbalanced, i.e., some categories happen more often than others, therefore matching to a larger category is more likely. Thus, the approaches perform correct retrieval for larger categories, which increases the total accuracy, but results in smaller mean accuracy.

\begin{figure}
	\centering
	\makebox{\parbox{0.95\linewidth}{ \centering
	\includegraphics[width=0.99\linewidth]{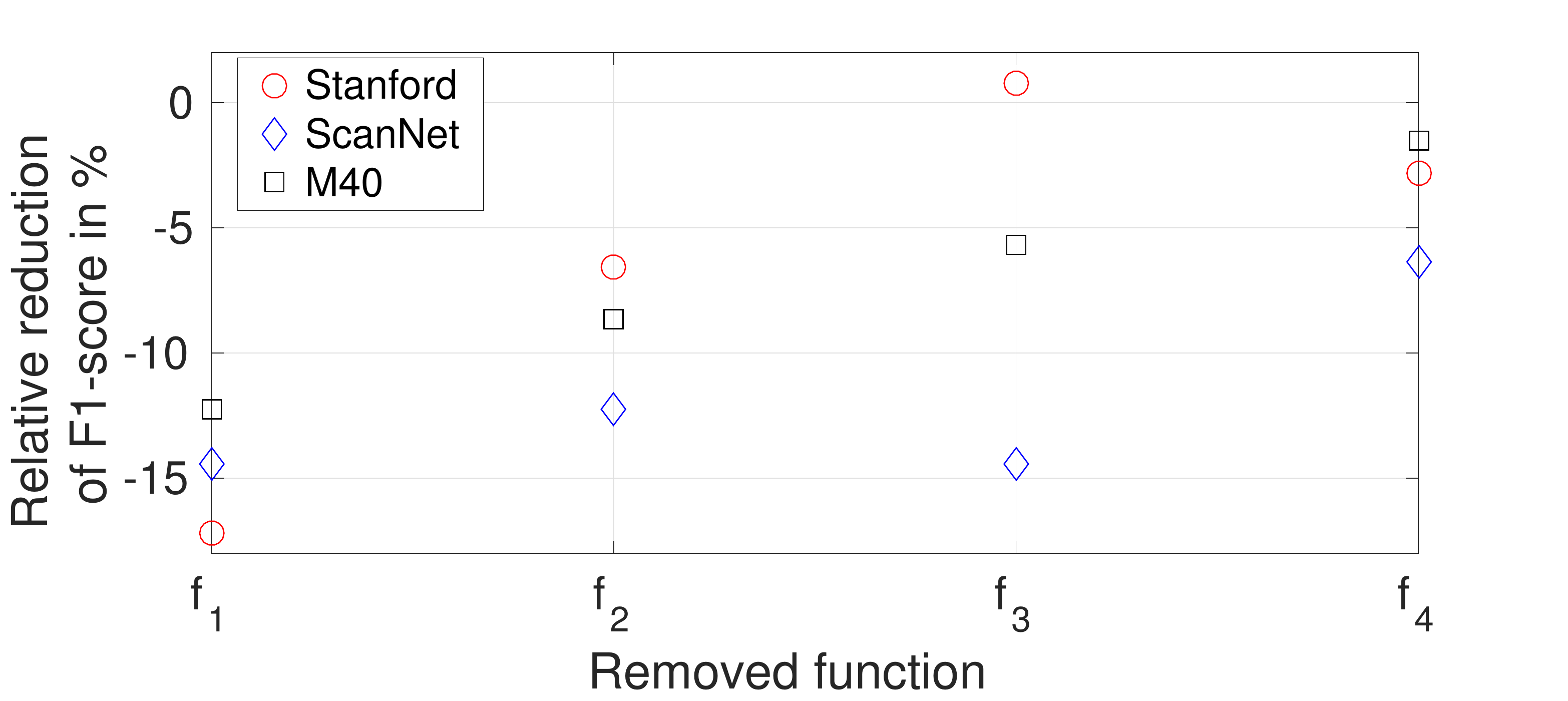}}}
	\caption{Illustration of the influence of function removal on the retrieval performance (F1-score) for the proposed descriptor. One function is removed at a time. Averaged over 10 runs.}
	\label{fig:eppf_study_dims}
\end{figure}

To gain further insights on the influence of various functions on the resulting performance, we disabled one PPF at a time and repeated retrieval experiments. The results for EPPF are given in Fig. {\ref{fig:eppf_study_dims}}. Here one can see that the largest drop in retrieval performance ($13-17\%$) is observed when the Euclidean distance feature is removed. The drop when removing the surface angle function is lower ($6-12\%$). Interestingly, the normal distance function performs differently on various datasets. In particular, on the Stanford dataset, which exhibits high levels of noise in normal orientation, removal of normal distance function leads to performance improvement by $1\%$. In contrast, on the other datasets, this does not happen and there is a significant drop by up to $15\%$. Finally, the visibility ratio function $f_4$ contributes the least to the overall performance on all datasets and results in a drop of $2-6\%$. This justifies the chosen number of bins for every dimension.

\subsection{Comparison of Deep Learning Approaches}

We further evaluate deep learning approaches on the task of object classification. We again use the Stanford, ScanNet and M40 datasets. We use the proposed 4D CNN network in combination with the handcrafted feature descriptor. For comparison, we also include 2D- and 3D-convolution-based networks (denotes as 2D and 3D, respectively). For optimization, we employ the Adam optimizer with a learning rate of $5\cdot10^{-4}$ and $0.5$ dropout probability. We perform training for $2,000$ epochs. Training on ScanNet takes 1-3 hours to converge with Tensorflow \cite{Tensorflow2016} and Nvidia Titan XP. For comparison, we also evaluate the method able to learn on point sets PointNet \cite{Qi2016}. We train the PointNet network on the given objects taking into account normalization into unit cube as advised by authors \cite{Qi2016}. We use standard parameters and feed the network with $2,048$ points. We perform training for $2,000$ epochs.

In Table~\ref{tab:deep_learning} we provide object classification results for both approaches (EPPF 4D denotes 4D convolutional network). One can observe that 4D convolution-based network performs better than the 2D- and 3D-variants. This is thanks to the fact that 4D co-occurrences between various dimensions are preserved. In contrast, by reshaping into 2D and 3D, such information is lost. On the Stanford and ScanNet datasets, 3D network performs better than the 2D-based one. One can further observe that our approach outperforms PointNet on the first two datasets. This can be explained by the fact that the proposed network can easily learn noise-resistant class-specific patterns based on handcrafted descriptors as compared to feeding the point sets directly. Notably, PointNet outperforms our approach on M40 dataset. Here, we obtain the PointNet result different from the one reported by authors in \cite{Qi2016} ($87.01\%$ vs. $89.2\%$), which is due to the fact that network training has random behaviour. With the lower level of noise in M40 dataset, PointNet is able to learn more descriptive representation for object classification. The lower performance of our network is due to the loss of information when operating on features instead of point sets. 

\begin{figure}
	\centering
	\makebox{\parbox{0.75\linewidth}{ \centering
	\includegraphics[width=0.95\linewidth]{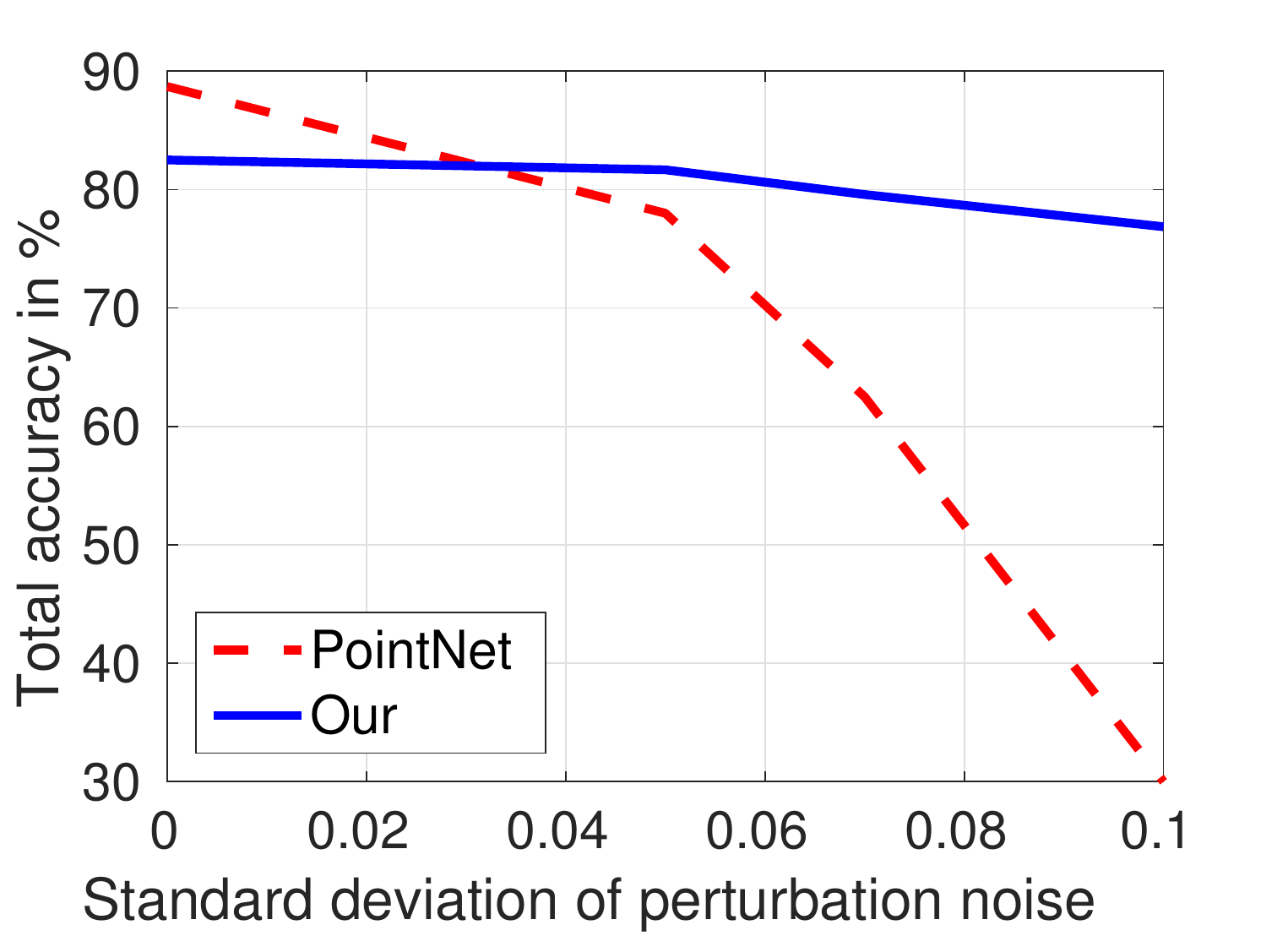}}}
	\caption{Illustration of the influence of zero-mean Gaussian random noise on the classification accuracy for the M40 dataset using 1024 points. The noise is added to each point independently. PointNet results from~\cite{Qi2016}.}
	\label{fig:m40_noise_results}
\end{figure}

To investigate the influence of noise on the total accuracy, we add zero-mean Gaussian random noise with various standard deviation values onto 3D coordinates of point sets. Then, we re-train the network using the noisy examples. The results for our 4D approach and PointNet are given in Fig.~\ref{fig:m40_noise_results}. Even though PointNet outperforms our approach on lower levels of noise, with increasing noise levels our approach suffers no significant decrease in accuracy. In contrast, PointNet performance starts to drastically deteriorate already at standard deviation values of $0.06$ (e.g., 6\% of the unit cube size). This can be explained by the fact that the proposed point pair functions are more robust to noise as compared to the network trained on point sets directly.

\begin{table*}[t]
\small
	\centering
	\caption{Classification performance of deep learning approaches using 2D, 3D and 4D convolutional layers.}
	\label{tab:deep_learning}
	\begin{tabular}{| c | c | c | c | c | c |}
	\hline 
	Dataset & Metric 	& PointNet \cite{Qi2016} 	& EPPF 2D & EPPF 3D & EPPF 4D  \\ \hline
	\multirow{4}{*}{Stanford} & Total accuracy (\%) 			& 64.30 & 82.01  & 81.94 & \textbf{83.22}   \\
	& Mean accuracy (\%) 	& 42.48 	& 64.26  & \textbf{66.37} & 65.11   \\
	& Mean recall (\%) 	& 40.47 	& 70.88 & 60.94 & \textbf{72.13}   \\
	& F1-score 		& 0.395 		& 0.652 & 0.665 & \textbf{0.672}   \\ 
    \hline
   	\multirow{4}{*}{ScanNet} & Total accuracy (\%)			& 63.04 	& 70.39 & 70.57 & \textbf{72.10}   \\
   	& Mean accuracy (\%)	& 37.50 	& 38.98 & 44.35 & \textbf{45.70}   \\
   	& Mean recall (\%)	& 19.53 	& \textbf{63.52} & 54.53 & 56.58   \\
   	& F1-score 		& 0.209 		& 0.433 & 0.472 & \textbf{0.488}   \\ 
    \hline 
    \multirow{4}{*}{M40} & Total accuracy (\%)	& \textbf{87.01} 	& 81.64 & 81.15 & 82.13   \\
       	& Mean accuracy (\%)	& \textbf{82.08} 	& 76.37 & 75.87 & 77.05   \\
       	& Mean recall (\%)	& \textbf{83.48} 	& 77.30 & 77.51 & 76.99   \\
       	& F1-score 		& \textbf{0.824} & 0.765 & 0.762 & 0.769   \\ 
    \hline
	\end{tabular}
\end{table*}

\section{Discussion}

\begin{figure}
\captionsetup[subfigure]{labelformat=empty}
\newcommand\m{0.33}
  \centering
  \makebox{\parbox{0.96\linewidth}{ \centering
  \subfloat{\includegraphics[width=\m\linewidth]{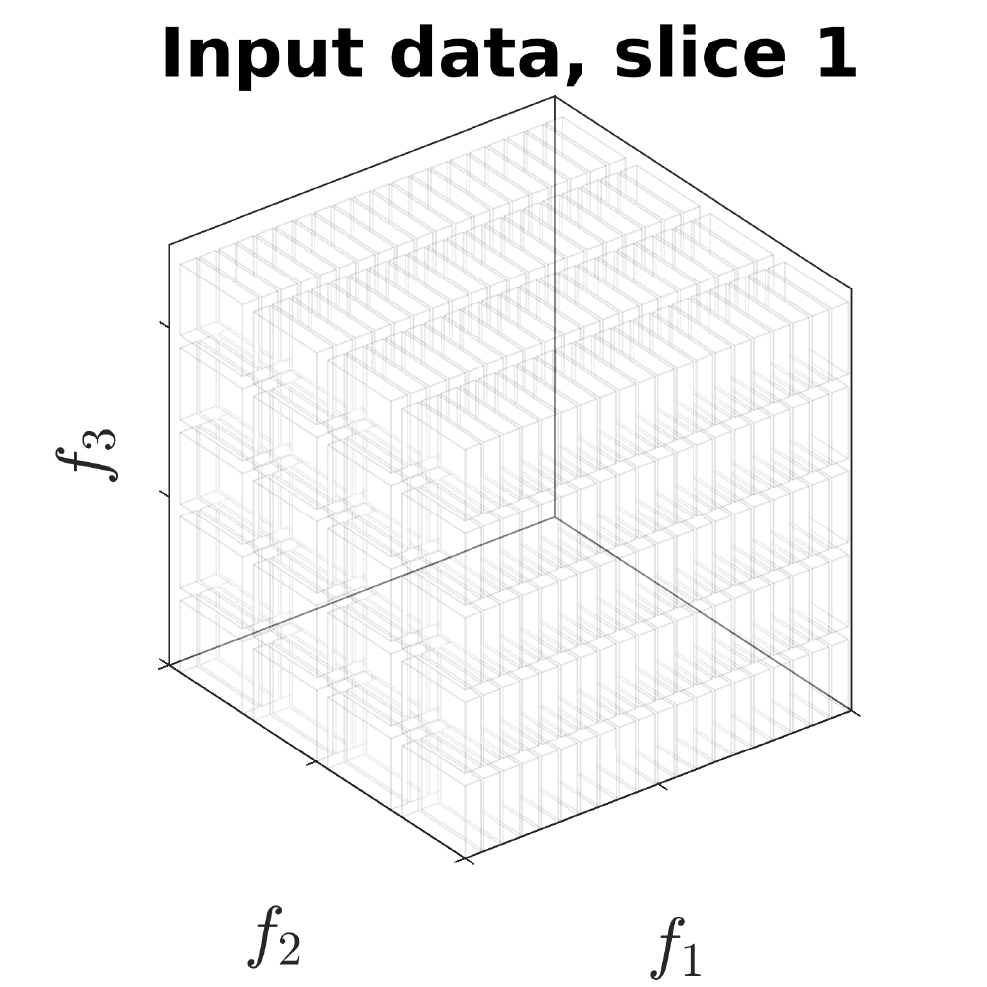}} \hfill
  \subfloat{\includegraphics[width=\m\linewidth]{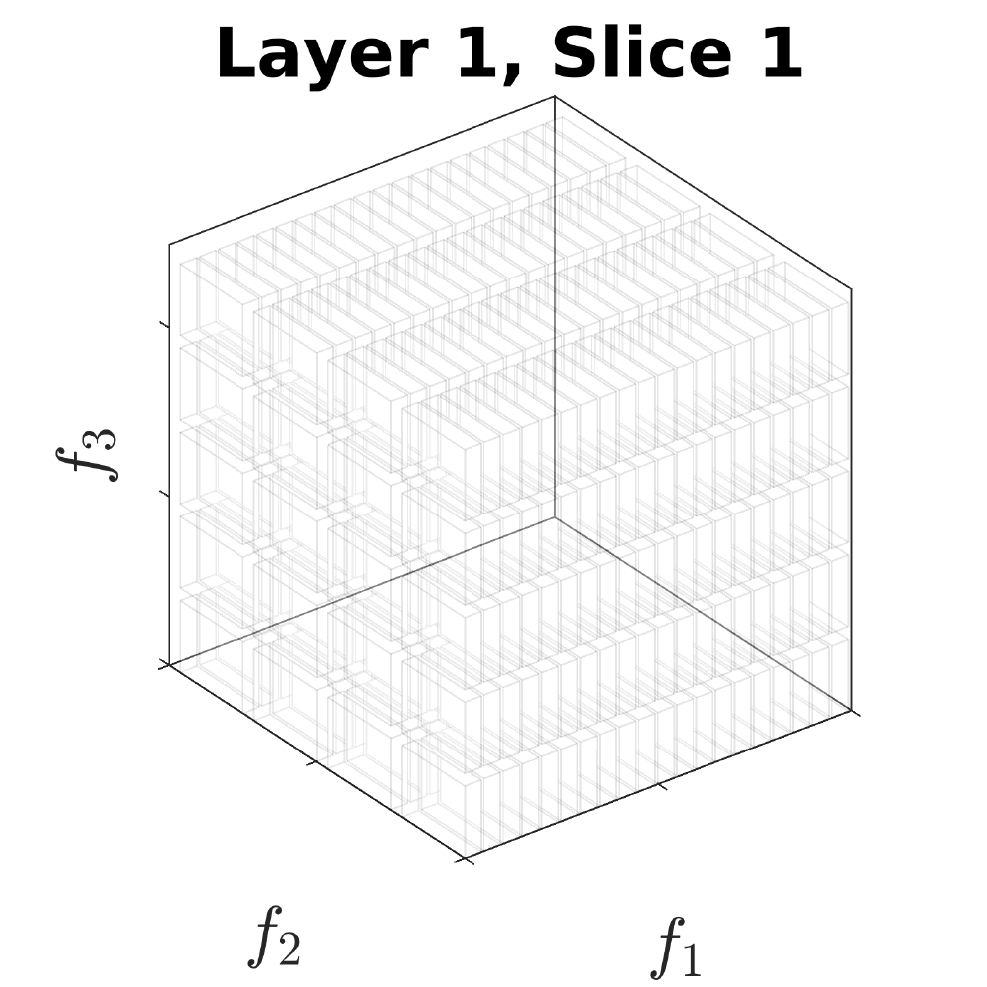}} \hfill
  \subfloat{\includegraphics[width=\m\linewidth]{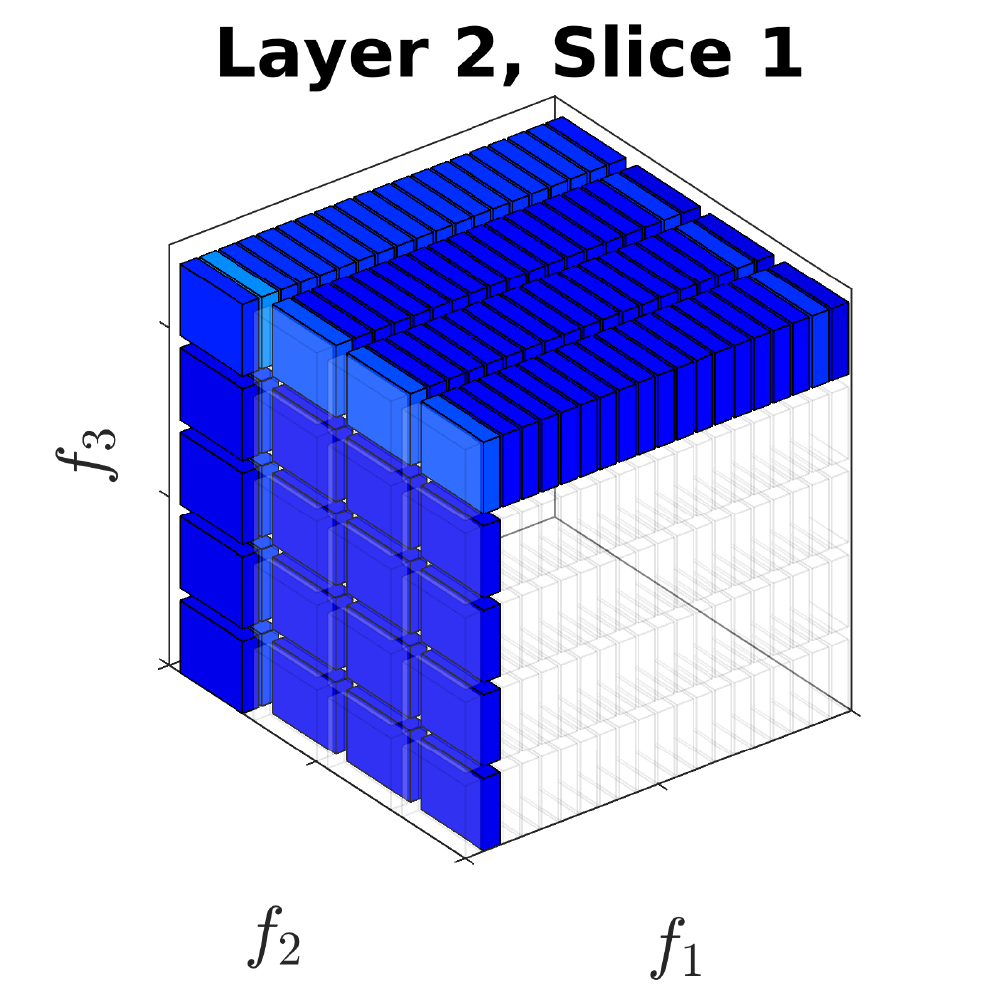}} \hfill
  \subfloat{\includegraphics[width=\m\linewidth]{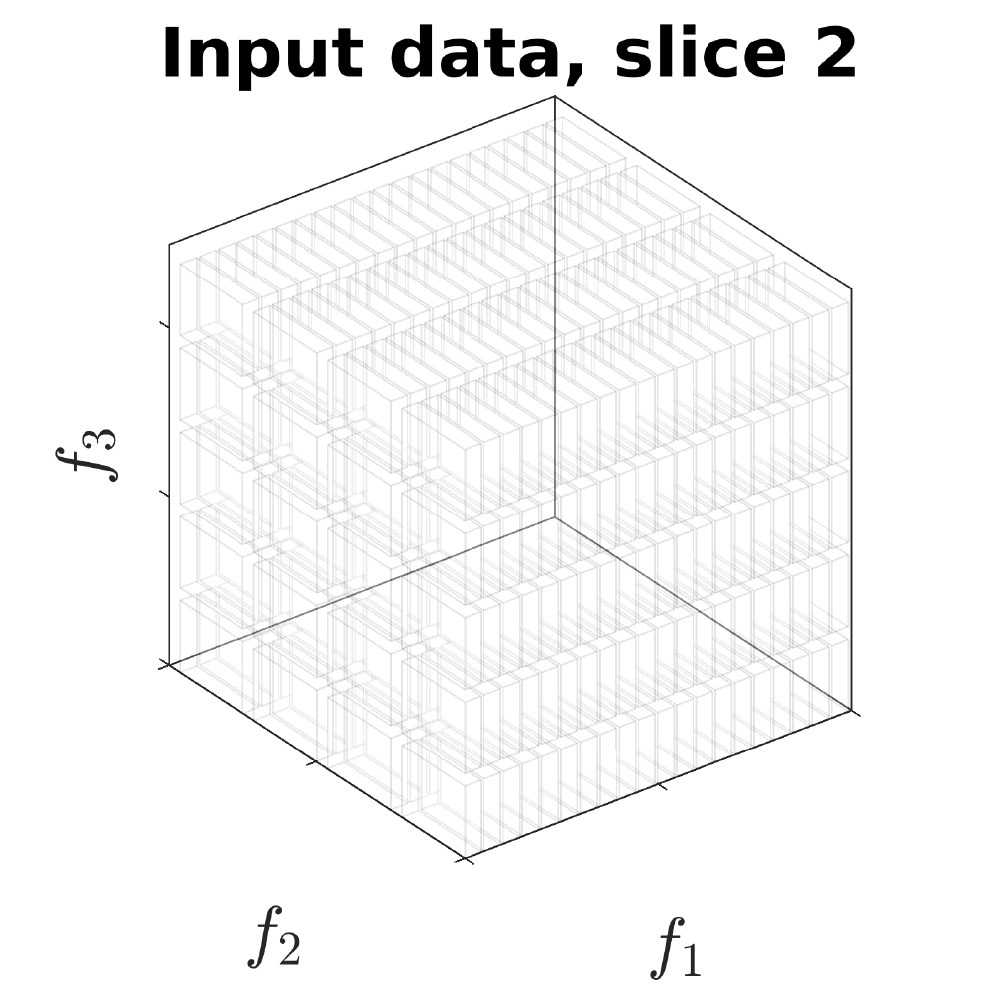}} \hfill
  \subfloat{\includegraphics[width=\m\linewidth]{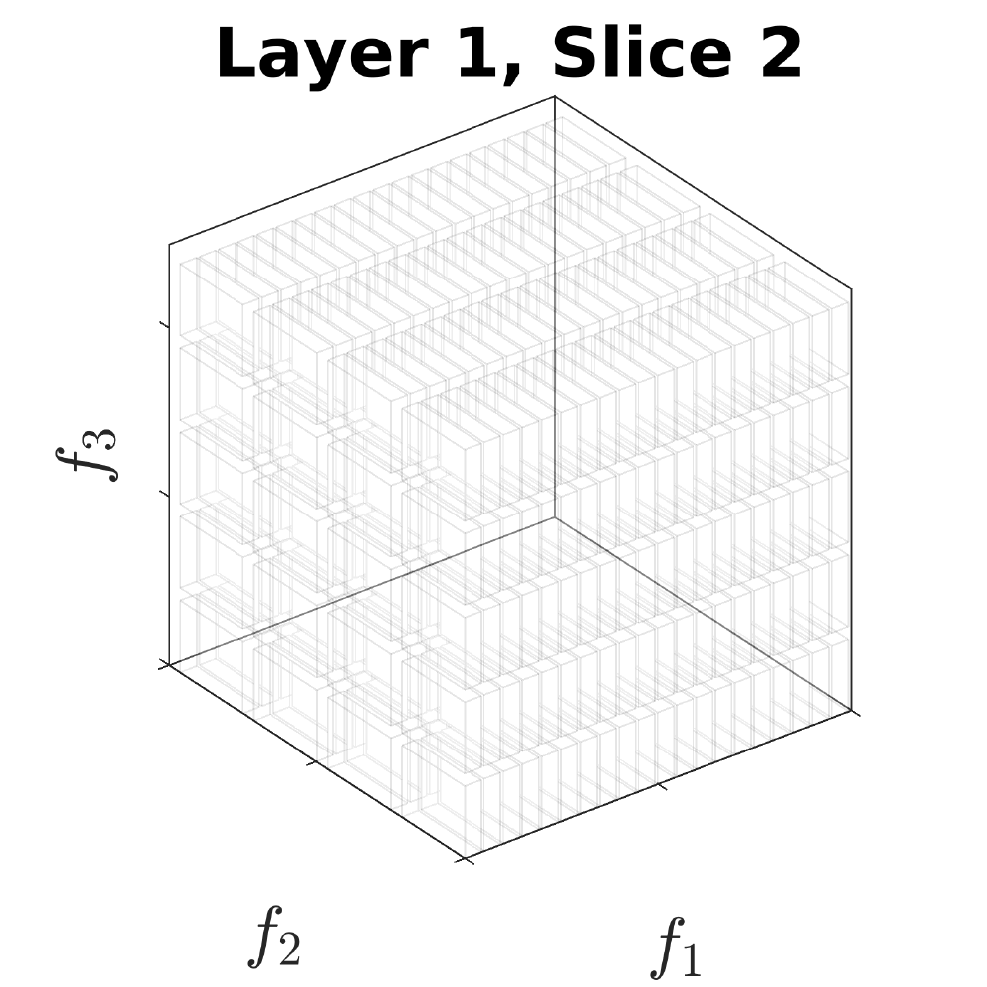}} \hfill
  \subfloat{\includegraphics[width=\m\linewidth]{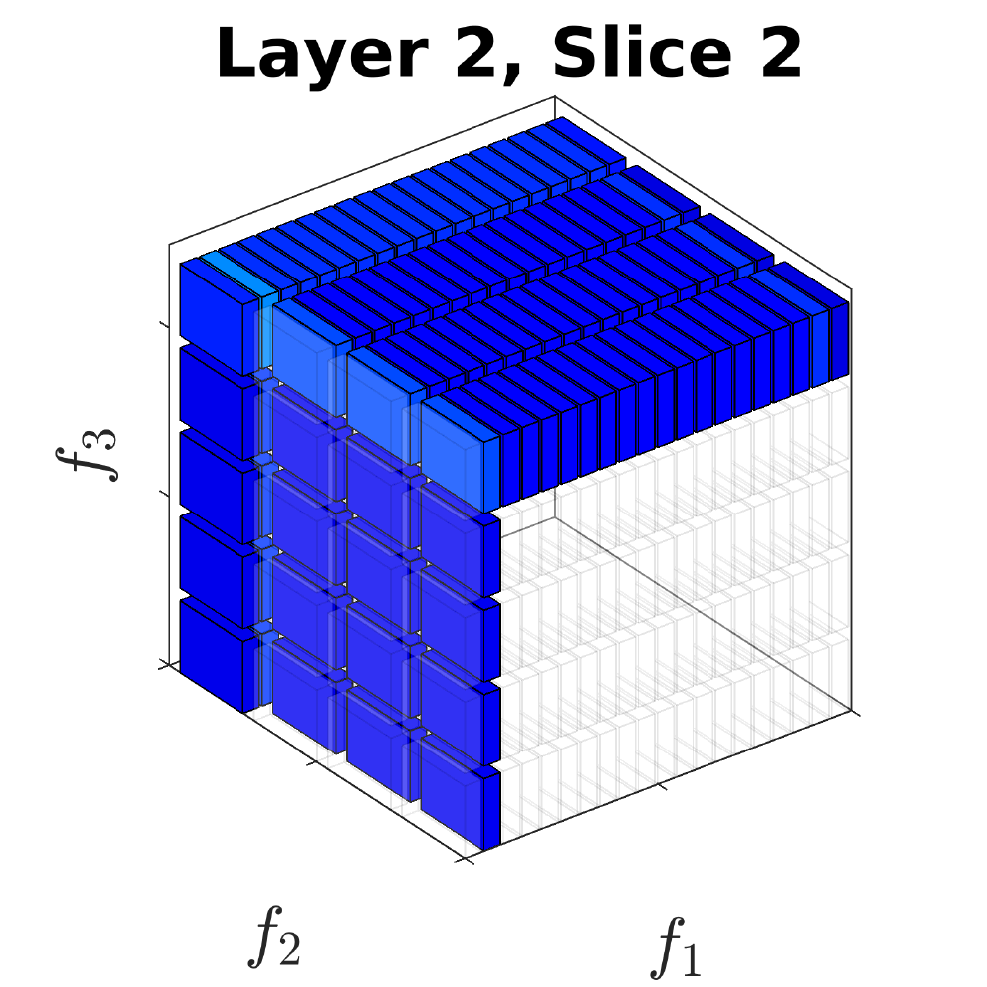}} \hfill
  \subfloat{\includegraphics[width=\m\linewidth]{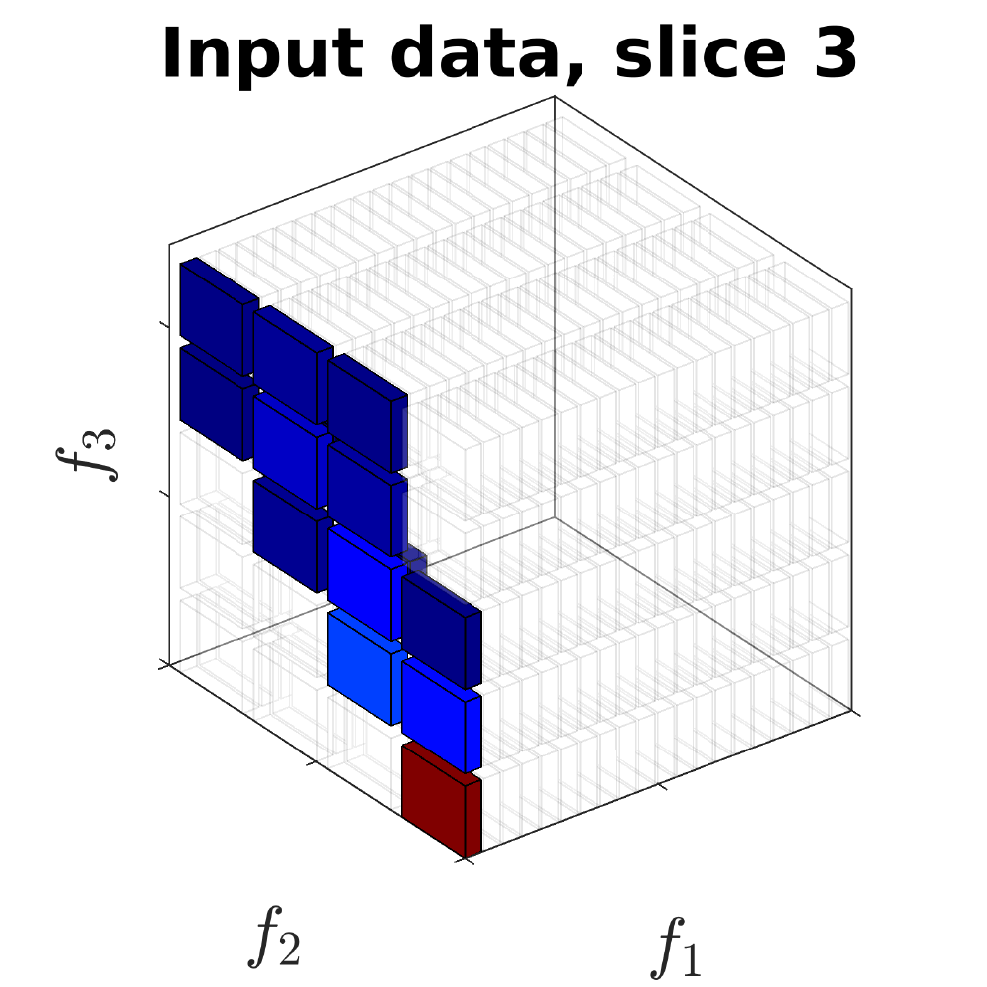}} \hfill
  \subfloat{\includegraphics[width=\m\linewidth]{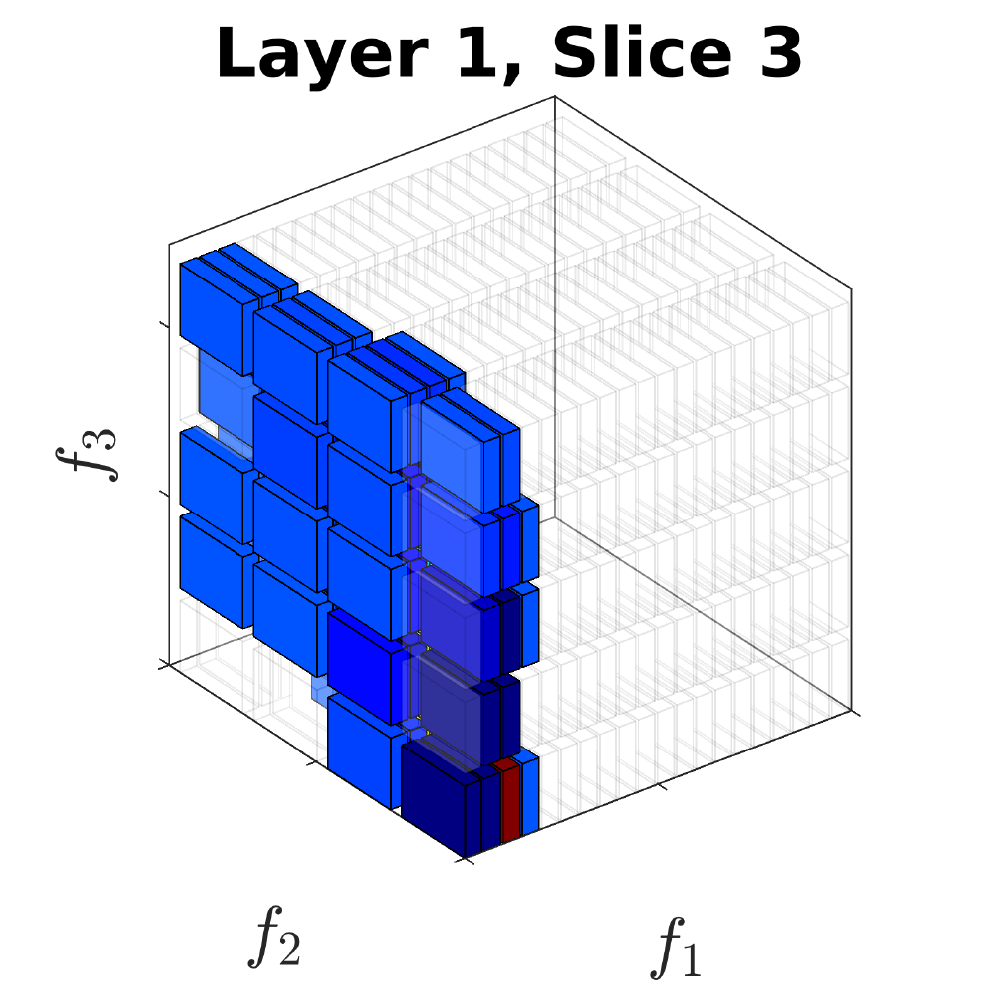}} \hfill
  \subfloat{\includegraphics[width=\m\linewidth]{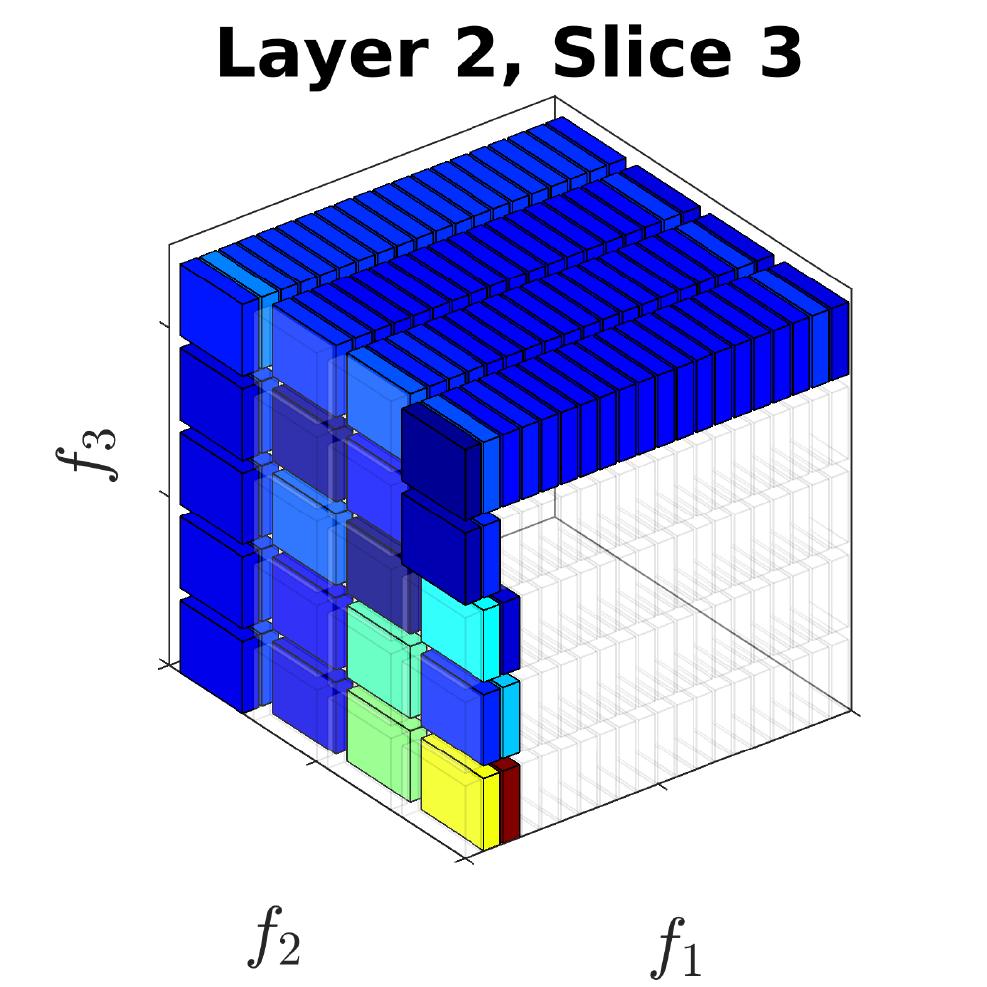}} \hfill
  }}
  \caption{Descriptor and 4D neural network responses for the object \textit{table} in the ScanNet dataset. Left: descriptor values. Middle: response of the first filter in the first layer. Right: filter response in the second layer. The rows show slices of the fourth dimension. Transparent bins correspond to constant offset values for the response (or 0 for the descriptor values), colored bins - to varying values. The bins are colored so that low values are shown in blue color, while high in red.}
\label{fig:network_response}
\end{figure}

\textbf{Network response visualization on different layers}. To gain further insights about the transformation learned by the network, we show its responses for an exemplary object. We choose the object \textit{table} in ScanNet and visualize the descriptor values and responses of the first filter in the first two layers in Fig.~\ref{fig:network_response}. Observe that the descriptor is very sparse, e.g., most part of the quantized space takes zero values. Curse of dimensionality is not a big issue here, as dimensionality of our function space is low (4D) and it is strongly quantized. We aggregate 20,000 4D function values into $1,200$ histogram bins. Hence, we get $16.67$ counts per bin on average, which further confirms that our space is sufficiently sampled. Furthermore, when feeding this descriptor into the first 4D convolutional layer, one can observe that the network has smeared this signal in the space. Finally, in the second layer the signal is even further spread across different dimensions. This is followed by a max-pooling layer that achieves invariance to spatial shift. The transformation learned by the network does not only perform simple Gaussian smoothing, but, more importantly, it amplifies the signal in certain regions and suppresses the signal in the other regions. This special perturbation benefits the generalization of our network, as the first 4D convolution layer can learn the fine features, which are characteristic for certain object categories, while suppressing occlusion and noise.

\textbf{Runtime analysis}. We review the runtime performance of the proposed descriptor. We implement the descriptor in C++ with OpenMP parallelization. We use Desktop PC Intel i7 with 24 GB RAM. The descriptor computation takes 8ms per object on average. This is comparable to runtime performance of the ESF, Wahl \etal and OUR-CVFH descriptors. This still allows us to use such descriptor in real-time operation in robotics for perception tasks. As our descriptor provides fixed feature size irrespective of the object dimensions, we expect relatively constant runtime when using the neural network for object classification.

\textbf{Further insights}. We have experimented with a number of network architectures for object classification. Nonetheless, we have not observed a significant improvement when using larger architectures, which can be explained by the limited size of the training data. Intuitively, reshaping operations performed in the proposed neural network removes information about the structure and feature co-occurrences. However, we have observed that 2D reshaping gives higher classification performance than using 4D blocks directly. This could be explained by the fact that the category-specific clusters learned by the network are spatially separated in all dimensions. We have alternatively considered a number of other strategies such as stacking the dimensions into a 2D representation, global max-pooling and did not observe any performance improvement.

PointNet generally took much longer to converge as compared to our approach on all datasets. For PointNet we also evaluated the voting scheme that applies multiple perturbations and uses majority vote of the representations as a prediction, but have observed no significant performance improvement. Furthermore, we have performed experiments with feeding point pairs directly to PointNet. This approach indeed slightly improved performance on noisy datasets, however, only by a small margin. To make sure no local optima influenced the evaluation, we trained several times and reported the best test accuracy. Most of the considered datasets have unbalanced categories, e.g., some categories (such as \textit{chair} in ScanNet) occur much more often than others (\textit{lamp}). This leads to the effect that the network is able to learn very complex patterns for often occurring objects, while it can only learn simple patterns for rarely occurring objects.

\textbf{Limitations}. Tuning of the hyper-parameters of the network could bring further improvements. In particular, one could expect that Res-Net structure could improve the result \cite{He2016}. Another important aspect is the point sampling strategy: we use a straightforward random sampling for this work. We expect that by performing non-random point sampling, one could improve the classification performance. For this, the techniques similar to the ones explained by Birdal and Ilic~\cite{Birdal2017} can be applied. Finally, end-to-end learning with the goal of identifying more descriptive point pair and point triplet functions could bring further improvements in classification performance.

\section{CONCLUSIONS}

We have proposed to feed the values of the global descriptor into the novel 4D neural network for object classification, which outperforms existing deep learning approaches on realistic data. We further verified that 4D convolutional layers outperform 2D and 3D convolutional layers. We have also illustrated that by carefully selecting the PPFs and the number of bins over different dimensions, one can enhance performance of the point pair-based global descriptor. Experimental results on 3 benchmark datasets confirm the superiority of such design in high noise and occlusion scenario. By providing a compact description as input data into a neural network one can make the learning problem easier and achieve faster convergence.

\bibliographystyle{IEEEtran}
\bibliography{IEEEabrv,ral_bib}

\begin{thebibliography}{10}
\providecommand{\url}[1]{#1}
\csname url@samestyle\endcsname
\providecommand{\newblock}{\relax}
\providecommand{\bibinfo}[2]{#2}
\providecommand{\BIBentrySTDinterwordspacing}{\spaceskip=0pt\relax}
\providecommand{\BIBentryALTinterwordstretchfactor}{4}
\providecommand{\BIBentryALTinterwordspacing}{\spaceskip=\fontdimen2\font plus
\BIBentryALTinterwordstretchfactor\fontdimen3\font minus
  \fontdimen4\font\relax}
\providecommand{\BIBforeignlanguage}[2]{{%
\expandafter\ifx\csname l@#1\endcsname\relax
\typeout{** WARNING: IEEEtran.bst: No hyphenation pattern has been}%
\typeout{** loaded for the language `#1'. Using the pattern for}%
\typeout{** the default language instead.}%
\else
\language=\csname l@#1\endcsname
\fi
#2}}
\providecommand{\BIBdecl}{\relax}
\BIBdecl

\bibitem{Birdal2015}
T.~Birdal and S.~Ilic, ``Point pair features based object detection and pose
  estimation revisited,'' in \emph{Proceedings of the International Conference
  on 3D Vision}, 2015, pp. 527--535.

\bibitem{Wohlkinger2011}
W.~Wohlkinger and M.~Vincze, ``Ensemble of shape functions for 3d object
  classification,'' in \emph{Proceedings of the IEEE International Conference
  on Robotics and Biomimetics}, Dec 2011, pp. 2987--2992.

\bibitem{Drost2010}
B.~Drost, M.~Ulrich, N.~Navab, and S.~Ilic, ``Model globally, match locally:
  Efficient and robust 3d object recognition,'' in \emph{Proceedings of the
  IEEE Conference on Computer Vision and Pattern Recognition (CVPR)}, June
  2010, pp. 998--1005.

\bibitem{Wahl2003}
E.~Wahl, U.~Hillenbrand, and G.~Hirzinger, ``Surflet-pair-relation histograms:
  a statistical 3d-shape representation for rapid classification,'' in
  \emph{Proceedings of IEEE International Conference on 3-D Digital Imaging and
  Modeling (3DIM)}, 2003, pp. 474--481.

\bibitem{Qi2016}
C.~R. Qi, H.~Su, K.~Mo, and L.~J. Guibas, ``Pointnet: Deep learning on point
  sets for 3d classification and segmentation,'' \emph{Proceedings of the IEEE
  Conference on Computer Vision and Pattern Recognition (CVPR)}, 2017.

\bibitem{Rusu2009}
R.~B. Rusu, N.~Blodow, and M.~Beetz, ``Fast point feature histograms (fpfh) for
  3d registration,'' in \emph{Proceedings of the IEEE International Conference
  on Robotics and Automation}, 2009, pp. 3212--3217.

\bibitem{Kasaei2016}
S.~H. Kasaei, L.~S. Lopes, A.~M. Tomé, and M.~Oliveira, ``An orthographic
  descriptor for 3d object learning and recognition,'' in \emph{Proceedings of
  the IEEE/RSJ International Conference on Intelligent Robots and Systems
  (IROS)}, Oct 2016, pp. 4158--4163.

\bibitem{Sanchez2015}
A.~J. Rodríguez-Sánchez, S.~Szedmak, and J.~Piater, ``Scurv: A 3d descriptor
  for object classification,'' in \emph{Proceedings of the IEEE/RSJ
  International Conference on Intelligent Robots and Systems (IROS)}, Sept
  2015, pp. 1320--1327.

\bibitem{Aldoma2012}
A.~Aldoma, F.~Tombari, R.~B. Rusu, and M.~Vincze, ``Our-cvfh -- oriented,
  unique and repeatable clustered viewpoint feature histogram for object
  recognition and 6dof pose estimation,'' \emph{Proceedings of the Joint 34th
  DAGM and 36th OAGM Symposium on Pattern Recognition}, pp. 113--122, 2012.

\bibitem{Furuya2015}
T.~Furuya and R.~Ohbuchi, ``Diffusion-on-manifold aggregation of local features
  for shape-based 3d model retrieval,'' in \emph{Proceedings of the 5th ACM
  International Conference on Multimedia Retrieval}, 2015, pp. 171--178.

\bibitem{Sedaghat2017}
N.~Sedaghat, M.~Zolfaghari, E.~Amiri, and T.~Brox, ``Orientation-boosted voxel
  nets for 3d object recognition,'' in \emph{Proceedings of the British Machine
  Vision Conference (BMVC)}, 2017.

\bibitem{Engelcke2017}
M.~Engelcke, D.~Rao, D.~Z. Wang, C.~H. Tong, and I.~Posner, ``Vote3deep: Fast
  object detection in 3d point clouds using efficient convolutional neural
  networks,'' \emph{Proceedings of the IEEE International Conference on
  Robotics and Automation (ICRA)}, pp. 1355--1361, 2017.

\bibitem{Li2016}
Y.~Li, S.~Pirk, H.~Su, C.~R. Qi, and L.~J. Guibas, ``Fpnn: Field probing neural
  networks for 3d data,'' in \emph{Advances in Neural Information Processing
  Systems (NIPS)}, 2016, pp. 307--315.

\bibitem{Su2015}
H.~Su, S.~Maji, E.~Kalogerakis, and E.~G. Learned{-}Miller, ``Multi-view
  convolutional neural networks for 3d shape recognition,'' in
  \emph{Proceedings of the IEEE International Conference on Computer Vision
  (ICCV)}, 2015.

\bibitem{Xie2016}
J.~Xie, G.~Dai, F.~Zhu, E.~K. Wong, and Y.~Fang, ``Deepshape: Deep-learned
  shape descriptor for 3d shape retrieval,'' \emph{IEEE Transactions on Pattern
  Analysis and Machine Intelligence (PAMI)}, vol.~39, no.~7, pp. 1335--1345,
  July 2016.

\bibitem{Ravanbakhsh2017}
S.~Ravanbakhsh, J.~Schneider, and B.~Poczos, ``Deep learning with sets and
  point clouds,'' in \emph{Proceedings of the International Conference on
  Learning Representations (ICLR) -- workshop track}, 2017.

\bibitem{Qi2017}
C.~R. Qi, L.~Yi, H.~Su, and L.~J. Guibas, ``Pointnet++: Deep hierarchical
  feature learning on point sets in a metric space,'' \emph{arXiv preprint
  arXiv:1706.02413}, 2017.

\bibitem{Hinterstoisser2016}
S.~Hinterstoisser, V.~Lepetit, N.~Rajkumar, and K.~Konolige, \emph{Going
  Further with Point Pair Features}.\hskip 1em plus 0.5em minus 0.4em\relax
  Springer International Publishing, 2016, pp. 834--848.

\bibitem{Bobkov2017}
D.~Bobkov, S.~Chen, M.~Kiechle, S.~Hilsenbeck, and E.~Steinbach,
  ``Noise-resistant unsupervised object segmentation in multi-view indoor point
  clouds,'' in \emph{Proceedings of the International Joint Conference on
  Computer Vision, Imaging and Computer Graphics Theory and Applications
  (VISIGRAPP)}, February 2017, pp. 149--156.

\bibitem{Wang2016}
T.-C. Wang, J.-Y. Zhu, E.~Hiroaki, M.~Chandraker, A.~A. Efros, and
  R.~Ramamoorthi, ``A 4d light-field dataset and cnn architectures for material
  recognition,'' in \emph{Proceedings of the European Conference on Computer
  Vision}.\hskip 1em plus 0.5em minus 0.4em\relax Springer, 2016, pp. 121--138.

\bibitem{Aldoma2012PCL}
A.~Aldoma, Z.-C. Marton, F.~Tombari, W.~Wohlkinger, C.~Potthast, B.~Zeisl,
  R.~B. Rusu, S.~Gedikli, and M.~Vincze, ``Tutorial: Point cloud library:
  Three-dimensional object recognition and 6 dof pose estimation,'' \emph{IEEE
  Robotics and Automation Magazine}, vol.~19, pp. 80--91, 2012.

\bibitem{Wu2015}
Z.~Wu, S.~Song, A.~Khosla, F.~Yu, L.~Zhang, X.~Tang, and J.~Xiao, ``3d
  shapenets: A deep representation for volumetric shapes,'' in
  \emph{Proceedings of the IEEE Conference on Computer Vision and Pattern
  Recognition (CVPR)}, 2015, pp. 1912--1920.

\bibitem{Armeni2016}
I.~Armeni, O.~Sener, A.~R. Zamir, H.~Jiang, I.~Brilakis, M.~Fischer, and
  S.~Savarese, ``3d semantic parsing of large-scale indoor spaces,'' in
  \emph{Proceedings of the IEEE Conference in Computer Vision and Pattern
  Recognition (CVPR)}, 2016.

\bibitem{Dai2017}
A.~Dai, A.~X. Chang, M.~Savva, M.~Halber, T.~Funkhouser, and M.~Nie{\ss}ner,
  ``Scannet: Richly-annotated 3d reconstructions of indoor scenes,'' in
  \emph{Proceedings of the IEEE Conference on Computer Vision and Pattern
  Recognition (CVPR)}, 2017.

\bibitem{Boulch2012}
A.~Boulch and R.~Marlet, ``Fast and robust normal estimation for point clouds
  with sharp features,'' \emph{Computer Graphics Forum}, vol.~31, no.~5, pp.
  1765--1774, Aug. 2012.

\bibitem{Tensorflow2016}
M.~Abadi, A.~Agarwal, P.~Barham, E.~Brevdo, Z.~Chen, C.~Citro, G.~S. Corrado,
  A.~Davis, J.~Dean, and M.~D. et~al., ``Tensorflow: Large-scale machine
  learning on heterogeneous distributed systems,'' \emph{arXiv preprint
  arXiv:1603.04467}, 2016.

\bibitem{He2016}
K.~He, X.~Zhang, S.~Ren, and J.~Sun, ``Deep residual learning for image
  recognition,'' in \emph{Proceedings of the IEEE Conference on Computer Vision
  and Pattern Recognition (CVPR)}, June 2016, pp. 770--778.

\bibitem{Birdal2017}
T.~Birdal and S.~Ilic, ``A point sampling algorithm for 3d matching of
  irregular geometries,'' in \emph{Proceedings of the IEEE/RSJ International
  Conference on Intelligent Robots and Systems (IROS)}, 2017.

\end{thebibliography}

\end{document}